\documentclass{clv2}

\expandafter\def\expandafter\normalsize\expandafter{%
    \normalsize
    \setlength\abovedisplayskip{2pt}
    \setlength\belowdisplayskip{2pt}
    \setlength\abovedisplayshortskip{0pt}
    \setlength\belowdisplayshortskip{0pt}
}

\usepackage{epsfig}

\usepackage{amsmath}
\usepackage{amssymb}
\usepackage{algorithm}
\usepackage{algpseudocode}

\usepackage{url}
\usepackage{multirow}

\usepackage{subcaption}
\usepackage{float}

\usepackage[inline]{enumitem}

\setlist[itemize]{itemsep=1pt, topsep=2pt}

\runningtitle{Meaning using Logical and Distributional Models}
\runningauthor{Beltagy, Roller, Cheng, Erk and Mooney}

\begin{document}

\title{Representing Meaning with a Combination of Logical and Distributional Models}

\author{I. Beltagy}
\affil{Computer Science Department\\  The University of Texas at Austin}
\author{Stephen Roller}
\affil{Computer Science Department\\  The University of Texas at Austin}
\author{Pengxiang Cheng}
\affil{Computer Science Department\\  The University of Texas at Austin}
\author{Katrin Erk}
\affil{Linguistics Department\\  The University of Texas at Austin}
\author{Raymond J. Mooney}
\affil{Computer Science Department\\  The University of Texas at Austin}

%\historydates{Submission received:            10th April 2015}
%\historydates{Revised version received:      5th May 2016}
%\historydates{Accepted for publication:        8th May 2016}

\maketitle

\begin{abstract}

NLP tasks differ in the semantic information they require, and at this
time no single semantic representation fulfills all
requirements. Logic-based representations characterize sentence
structure, but do not capture the graded aspect of
meaning. Distributional models give graded similarity ratings for
words and phrases, but do not capture sentence structure in the same
detail as logic-based approaches. So it has been argued that the two are complementary.

We adopt a hybrid approach that combines logical and
distributional semantics using probabilistic logic, specifically
Markov Logic Networks (MLNs). In this paper, we focus on the three components
of a practical system:\footnote{System is available for download at: \url{https://github.com/ibeltagy/pl-semantics}}
%integrating logical and distributional models
1) \emph{Logical representation} focuses on representing the input problems 
in probabilistic logic. 
%This is quite different from representing them in standard first-order logic. 
2) \emph{Knowledge
base construction} creates weighted inference rules by integrating
distributional information with other sources.
% , notably WordNet and
% an existing paraphrase collection. In particular, we use our system to
% evaluate distributional lexical entailment approaches. We use a logic-based 
% approach  to determine the necessary inference rules. 
% %More sources can easily be added by mapping them to logical rules; 
% Our system learns a resource-specific weight that corrects for scaling differences
% between resources.
3) \emph{Probabilistic inference} involves
solving the resulting MLN inference problems efficiently. 
To evaluate our approach, we use the task of textual entailment (RTE),
which can utilize the strengths of both logic-based and distributional
representations. In particular we focus on the SICK dataset, where we
achieve state-of-the-art results.
We also release a lexical entailment dataset of 10,213 rules extracted from the 
SICK dataset, which is a valuable resource for evaluating lexical entailment systems.\footnote{Available at: \url{https://github.com/ibeltagy/rrr}}

\end{abstract}

\section{Introduction}
\label{sec:intro}
Computational semantics studies mechanisms for encoding the meaning of natural language
in a machine-friendly representation that supports automated reasoning and
that, ideally, can be automatically acquired from large text corpora.
Effective semantic representations and reasoning
tools give computers the power to perform complex applications like
question answering. But
applications of computational semantics are very diverse and pose differing requirements on
the underlying representational formalism. Some applications benefit
from a detailed representation of the structure of complex sentences. Some applications
require the ability to recognize near-paraphrases or degrees of similarity
between sentences. Some applications require inference, either exact or
approximate. 
Often it is necessary to handle ambiguity and vagueness in meaning. 
Finally, we frequently want to learn knowledge
relevant to these applications automatically from corpus data.

There is no single representation for natural language meaning at this
time that fulfills all of the above requirements, but there are representations
that fulfill some of them.  Logic-based
representations~\cite{montague:theoria70,dowty:book81,kamp:book93} %\nocite{thomason:book74}
like first-order logic represent many linguistic phenomena like negation, quantifiers, or
discourse entities. 
Some of these phenomena (especially negation scope and discourse entities over paragraphs) 
can not be easily represented in syntax-based representations like Natural Logic~\cite{maccartney:iwcs09}.
In addition, first-order
logic has standardized inference mechanisms. 
Consequently, logical approaches have been widely used in semantic parsing where it
supports answering complex natural language queries requiring reasoning and data aggregation
\cite{zelle:aaai96,kwiatkowski:emnlp13,pasupat:acl15}.
But logic-based representations often rely on manually constructed
dictionaries for lexical semantics, which can result in coverage
problems. And first-order logic, being binary in nature, does not
capture the graded aspect of meaning (although there are
combinations of logic and probabilities). 
% KE: I'd rather not say this because it misses the point: We can
% combine probabilities and first-order logic, with and without
% distributional modeling. 
Distributional models~\cite{turney:jair10} use
contextual similarity to predict the graded semantic similarity of words and
phrases~\cite{landauer:psychrev97,mitchell:jcs10},
and to model
polysemy~\cite{schutze:cl98,erk:emnlp08,thater:acl10}. But at this
point, fully representing structure and logical form using distributional models of
phrases and sentences is still an open problem. Also, current distributional representations do not support logical inference
that captures the semantics of negation, logical connectives, and quantifiers.
Therefore, distributional models and logical
representations of natural language meaning are complementary in their
strengths, as has frequently been remarked~\cite{ClarkCoeckeSadr,garrette:iwcs11,grefenstette:emnlp11,Baroni:2012wr}.

Our aim has been to construct a general-purpose natural language
understanding system that provides in-depth representations of
sentence meaning amenable to automated inference, but that also
allows for flexible and graded inferences involving word meaning. 
Therefore, our approach combines logical and distributional
methods. Specifically, we use first-order logic as a basic representation,
providing a sentence representation that can be easily
interpreted and manipulated. However, we also use distributional
information for a more graded representation 
of words and short phrases, providing information on
near-synonymy and lexical entailment.  Uncertainty and gradedness at the lexical and
phrasal level should inform inference at all levels, so we rely on
probabilistic inference to integrate logical and distributional
semantics. Thus, our system has three main
components, all of which present interesting challenges. For logic-based
semantics, one of the  challenges is to adapt the representation to the
assumptions of the probabilistic logic~\cite{Beltagy:IWCS15}. For
distributional lexical and phrasal semantics, one challenge is to
obtain appropriate weights for inference rules~\cite{roller:coling14}. 
In probabilistic inference, the core challenge is formulating the problems to allow
for efficient MLN inference ~\cite{beltagy:starai14}.

Our approach has previously been described in \namecite{garrette:iwcs11} and
\namecite{beltagy:starsem13}. We have demonstrated the generality of the system by
applying it to both textual entailment (RTE-1 in
\namecite{beltagy:starsem13}, SICK (preliminary results) and FraCas in
\namecite{Beltagy:IWCS15}) and semantic textual similarity (STS)
similarity~\cite{beltagy:acl14}, and we are investigating applications to
question answering. We have demonstrated the
modularity of the system by testing both Markov Logic Networks~\cite{richardson:mlj06} and
Probabilistic Soft Logic~\cite{broecheler:uai10} as probabilistic
inference engines~\cite{beltagy:starsem13,beltagy:acl14}. 

The primary aim of the current paper is to describe our complete system in detail, all the
nuts and bolts necessary to bring together the three distinct
components of our approach, and to showcase some of the difficult
problems that we face in all three areas along with our current solutions.

The secondary aim of this paper is to show that it is possible to take
this general approach and apply it
to a specific task, here textual entailment~\cite{dagan:hlt13}, adding
task-specific aspects to the general framework in such a way
that the model achieves
state-of-the-art performance. We chose the task of textual entailment 
because it utilizes the strengths of both logical and
distributional representations. We specifically use the SICK
dataset~\cite{marelli:lrec14} because it was designed to focus on lexical knowledge
rather than world knowledge, matching the focus of our system. 

Our system is flexible with respect to the sources of lexical and
phrasal knowledge it uses, and in this paper
we utilize PPDB~\cite{ganitkevitch:naacl13} and WordNet along with
distributional models. But we are specifically interested in
distributional models, in particular in how well they can predict
lexical and phrasal entailment. 
Our system provides a unique framework
for evaluating distributional models on RTE because the overall sentence representation is handled by the
logic, so we can zoom in on the performance of distributional models at
predicting lexical ~\cite{Geffet:2005tp} and phrasal entailment. The evaluation of
distributional models on RTE is the third aim of our paper. 
%Aims two and three are somewhat at odds in that
%to achieve aim two, w
We build a lexical entailment classifier that exploits both
task-specific features as well as
distributional information, and present an in-depth
evaluation of the distributional components.

We now provide a brief sketch of our
framework~\cite{garrette:iwcs11,beltagy:starsem13}. Our framework is three components, 
the first is the logical form which is the primary meaning representation for a sentence.
The second is the distributional information which is encoded in the form of \emph{weighted}
logical rules (first-order formulas). For example, in its simplest form, our approach
can use the distributional similarity of the words
\textit{grumpy} and \textit{sad} as the weight on a rule that says
if $x$ is grumpy then there is a chance that $x$ is also sad:
\[\forall x.grumpy(x) \to sad(x) \mid f(sim(\vec{grumpy},
\vec{sad}))\]
where $\vec{grumpy}$ and $\vec{sad}$ are the vector representations of the words \textit{grumpy} and \textit{sad}, 
$sim$ is a distributional similarity measure, like cosine, and $f$ is a function that maps the similarity score to an MLN weight.
A more principled, and in fact superior, choice is to use an
asymmetric similarity measure to compute the weight, as we discuss
below.

The third component is inference. We draw inferences over the weighted rules using Markov Logic Networks
(MLN)~\cite{richardson:mlj06}, a Statistical Relational Learning (SRL)
technique~\cite{getoor:book07} that combines logical and statistical knowledge
in one uniform framework, and provides a mechanism for coherent probabilistic
inference.  MLNs represent uncertainty in terms of weights on
the logical rules as in the example below:
\begin{equation}
	\begin{aligned}
\forall & x.  \: ogre(x) \Rightarrow grumpy(x)  \: | \:  1.5 \\
%\forall & x.y \: friend(x,y) \Rightarrow ( ogre(x) \Leftrightarrow ogre(y))  \: | \:  1.1
\forall & x,y. \: ( friend(x,y) \wedge ogre(x) ) \Rightarrow  ogre(y)  \: | \:  1.1
	\end{aligned}
\label{eq:smoke}
\end{equation}
which states that there is a chance that ogres are grumpy, 
and friends of ogres tend to be ogres too. Markov logic
uses such weighted rules to derive a probability distribution
over possible worlds through an undirected graphical model. This
probability distribution over possible worlds is then used to draw inferences.

We publish a dataset of the lexical and phrasal  
rules that our system queries when running on SICK, 
along with gold standard annotations.
The training and testing sets are extracted 
from the SICK training and testing sets respectively. The total number of rules 
(training + testing) is 12,510, only 10,213 are unique with 3,106 entailing rules, 
177 contradictions and 6,928 neutral. 
This is a valuable resource for testing lexical entailment systems, 
containing a variety of entailment relations (hypernymy, synonymy, antonymy,
etc.) that are actually useful in an end-to-end RTE system.

%%%%%% contributions
In addition to providing further details on the approach introduced in
\namecite{garrette:iwcs11} and \namecite{beltagy:starsem13} (including
improvements that improve the scalability of MLN
inference~\cite{beltagy:starai14} and adapt logical constructs for
probabilistic inference~\cite{Beltagy:IWCS15}) this paper makes the following
new contributions:
\begin{itemize}[leftmargin=9pt]

\item We show how to represent the  RTE task as an
inference problem in  probabilistic logic (sections~\ref{sec:hGivenT}, ~\ref{sec:dca}),
arguing for the use of a closed-word assumption (section~\ref{sec:cwa}).
%and show how to implement it for different forms of the hypothesis. 
%%%We argue that the closed-world assumption is a better fit for the RTE task
%%%because it matches the task definition, makes results less sensitive to 
%%%the domain size (number of constants in the domain), and enables
%%%inference optimization. 

\item Contradictory RTE sentence pairs are often only contradictory
  given some assumption about entity coreference. 
  For example,  \textit{An ogre is not snoring} and \textit{An ogre is snoring}
  are not contradictory unless we assume that the two ogres
  are the same. Handling such coreferences is important to detecting
  many cases of contradiction (section~\ref{sec:coref}). 
 
%\item To reduce the impact of misparsing, we combine results from two
%  different CCG parsers. Experiments show that this improves accuracy.
\item We use multiple parses to reduce the impact of misparsing (section~\ref{sec:boxer}).

\item In addition to distributional rules, we add rules from existing
  databases, in particular WordNet~\cite{princeton:url10} and the
  paraphrase collection PPDB~\cite{ganitkevitch:naacl13} (section~\ref{sec:precompiled}). 
%We use a rule-based 
%technique to translate entries from the paraphrase collection to
%logical rules. 

%\item Previously, our system did not use any lexical alignment of the Text
%  and the Hypothesis in the RTE task, but instead generated distributional
%  inference rules linking {\it any} word in the Text to {\it any} word in the
%  Hypothesis. We now use the logical form to guide alignment through a
%  variant of Robinson resolution~\cite{robinson:jacm65}, such that 
%  the only distributional rules constructed are those that are needed
%  for a successful inference. 
%  These rules can be annotated almost automatically with gold standard annotations.
\item  A logic-based alignment to guide generation of distributional rules (section~\ref{sec:rr}).

%We publish a dataset of all the lexical and phrasal distributional 
%rules collected from the SICK dataset using our variant of the
%Robinson resolution algorithm (12,510 rules), 
%along with gold standard annotations of entailment or non-entailment. 
%This is a valuable resource especially for testing lexical entailment systems, 
%as they contain a variety of entailment relations, 
%and are actually useful in an end-to-end RTE system.
\item  A dataset of all lexical and phrasal rules needed for the SICK dataset (10,213 rules). 
This is a valuable resource for testing lexical entailment systems 
on entailment relations that are actually useful in an end-to-end RTE system (section~\ref{sec:annotaterules}). 

%\item We also evaluate a compositional distributional approach on the task of
%phrasal entailment. Compositional distributional approaches have typically been
%  evaluated on tasks of phrase or sentence
%  similarity~\cite{mitchell:jcs10,paperno:acl14}. To our knowledge, it
%  has not been tested before to what 
%extent phrase similarity can be effectively used to determine entailment. The
%approach that we test is the 
%  state-of-the-art approach by \namecite{paperno:acl14}. We find that
%  this approach is effective at flagging phrase pairs that are
%  \emph{not} entailing, for example because of prepositions that
%  change sentence meaning or because of a difference in semantic roles
%  (``man eats near kitten''/''kitten eats''), but not so much at
%  identifying entailing phrase pairs. 
%
%  % Our experiments on lexical and phrasal entailments rely on the Robinson resolution based alignment, which removes
%  % large amounts of irrelevant rules and allows the system to focus on
%  % the relevant ones.
\item Evaluate  a state-of-the-art compositional distributional approach ~\cite{paperno:acl14}
on the task of phrasal entailment (section~\ref{sec:pengxiang}).  

%\item Rules from different sources come with different weights. We use 
%weight learning to map these weights to MLN weights. We learn one 
%weight scaling factor per rules source. We use simple grid-search 
%to learn the scaling factors.
\item A simple weight learning approach to map rule weights to MLN weights (section~\ref{sec:wlearn}). 

%\item Lexical entailment was defined by \namecite{Geffet:2005tp}
%as a relation that holds between two words if ``there
%are some contexts in which one of the words can be substituted by the
%other, such that the meaning of the original word can be inferred from
%the new one.'' At this point, it is unclear to what extent distributional
%information actually contains the information needed for lexical entailment. A recent
%paper title asks: ``Do supervised distributional methods really learn
%lexical inference relations?''~\cite{levy:2015tt}. We test this for
%the case of word pairs in the SICK dataset. Previous datasets
%for this task came from a variety of sources; we perform this task
%for the first time on data from an actual RTE
%dataset. Testing the ability of distributional similarity ratings on
%their ability to indicate lexical entailment, we confirm that
%distributional information contains some information on lexical entailment,
%though high cosine similarity often indicates co-hyponymy, and that
%the difference between two vectors is a better indicator of hypernymy
%than cosine, or than the supposed hypernym vector alone. 
%% We also find
%% that cosine similarity can identify some context-specific entailments
%% that WordNet does not pick up.
\item The question 
``Do supervised distributional methods really learn
lexical inference relations?''~\cite{levy:2015tt} has been studied 
before on a variety of lexical entailment datasets. 
For the first time, we study it on data from an actual RTE dataset
and show that distributional information is useful for lexical entailment (section~\ref{sec:lexeval}). 
%though high cosine similarity often indicates co-hyponymy, and that
%the difference between two vectors is a better indicator of hypernymy
%than cosine, or than the supposed hypernym vector alone. 

\item \namecite{marelli:semeval14} report that for the SICK dataset used in SemEval 2014, 
the best result was achieved by 
%what they call non-compositional systems -- 
systems that did not compute a sentence representation in a compositional manner. 
%In this paper, we show that it is
%possible for a model that performs deep compositional semantic analysis to reach
%state-of-the-art performance.
We present a model that performs deep compositional semantic analysis and achieves
state-of-the-art performance (section~\ref{sec:rteeval}). 

\end{itemize}

% The rest of this paper is organized as follows. Section \ref{sec:relwork} 
% provides somebackground. Section \ref{sec:overview} gives an
% overview of our system, with details in Sections
% \ref{sec:parse}, \ref{sec:kb} and \ref{sec:infer}. Section
% \ref{sec:eval} reports an evaluation on the SICK dataset, 
% and Section \ref{sec:future} discusses future
% work.

\section{Background}
\label{sec:relwork}
\paragraph{Logical Semantics}
Logical representations of meaning 
have a long tradition in linguistic
semantics~\cite{montague:theoria70,dowty:book81,kamp:book93,alshawi:92} and
computational semantics~\cite{blackburn:book05,vanEijckUnger}, 
and commonly used in semantic parsing~\cite{zelle:aaai96,berant:emnlp13,kwiatkowski:emnlp13}.
They handle many
complex semantic phenomena such as negation and quantifiers, they
identify discourse referents along with the predicates that apply to them and
the relations that hold between them. However, standard first-order logic and theorem provers  
are binary in nature, which prevents them from capturing the
graded aspects of meaning in language: Synonymy seems to come in
degrees~\cite{Edmonds:2000vc}, as does the difference between senses
in polysemous
words~\cite{brown:2008:ACLShort}. \namecite{vanEijckLappin} write: ``The case for abandoning the categorical view of competence and
adopting a probabilistic model is at least as strong in semantics as it is in syntax.''

Recent
wide-coverage tools that use logic-based sentence representations
include \namecite{CopestakeFlickinger}, \namecite{bos:step08}, 
and \namecite{lewis:tacl13}.  We use Boxer~\cite{bos:step08},  a wide-coverage 
semantic analysis tool that produces logical forms using Discourse
Representation Structures~\cite{kamp:book93}. It builds on the 
C\&C CCG (Combinatory Categorial Grammar) parser~\cite{clark:acl04}
and maps sentences into a lexically-based logical form, in
which the predicates are mostly words in the sentence. 
For example, the sentence \textit{An ogre loves a princess} is mapped to:
\begin{equation}
\exists x, y, z. \: ogre(x) \wedge agent(y, x) \wedge love(y) \wedge patient(y, z) \wedge princess(z)
\label{eq:boxer}
\end{equation}
As can be seen, Boxer uses a neo-Davidsonian framework~\cite{NeoDavidsonian}: $y$ is an
event variable, and the semantic roles $agent$ and $patient$ are
turned into predicates linking $y$ to the agent $x$ and patient $z$. 

As we discuss below, we combine Boxer's logical form
with weighted rules and perform probabilistic
inference. \namecite{lewis:tacl13} also integrate logical and
distributional approaches, but use distributional information to create
predicates for a standard binary logic and do not use probabilistic inference.
Much earlier, \namecite{hobbs:acl88} combined logical form
with weights in an abductive framework. There, the aim was to model the interpretation
of a passage as its best possible explanation.

\paragraph{Distributional Semantics}
Distributional models~\cite{turney:jair10} use statistics
on contextual data from large corpora to predict semantic similarity of words
and phrases~\cite{landauer:psychrev97,mitchell:jcs10}.
They are motivated by the observation that semantically
similar words occur in similar contexts, so words can be represented as vectors
in high dimensional spaces generated from the contexts in which they
occur~\cite{landauer:psychrev97,lund:behavior96}.  Therefore, 
distributional models are relatively easier to build than logical 
representations, automatically acquire knowledge from ``big data'', 
and capture the \emph{graded} nature of linguistic meaning, but they do 
not adequately capture logical
structure~\cite{grefenstette:starsem13}.   

Distributional models have also been
extended to compute vector representations for larger phrases, e.g. by adding
the vectors for the individual words~\cite{landauer:psychrev97} or by a
component-wise product of word
vectors~\cite{mitchell:acl08,mitchell:jcs10}, or through more complex methods
that compute phrase vectors from word vectors and
tensors~\cite{baroni:emnlp10,grefenstette:emnlp11}. 

\paragraph{Integrating logic-based and distributional semantics} 
It does not seem particularly useful at this point to speculate about phenomena
that either a distributional approach or a logic-based approach would
not be able to handle in principle, as both frameworks are continually
evolving. 
However, logical and distributional approaches clearly
differ in the strengths that they currently possess~\cite{ClarkCoeckeSadr,garrette:iwcs11,Baroni:2012wr}.
Logical form excels at in-depth representations of sentence structure and 
provides an explicit representation of
discourse referents. Distributional approaches are particularly good at representing
the meaning of words and short phrases in a way that allows for modeling degrees of similarity
and entailment and for modeling word meaning in context.  
This  suggests that it may be useful to combine the two frameworks. 

Another argument for combining both representations is that 
it makes sense from a theoretical point of view to address
meaning, a complex and multifaceted phenomenon, through a combination
of representations. Meaning is about truth, and
logical approaches with a model-theoretic semantics nicely address
this facet of meaning. Meaning is also about a community of speakers
and how they use language, and distributional models 
aggregate observed uses from many speakers.
%To combine logical and distributional information, we opt for
%probabilistic inference, as we think that gradience is so
%central to the representation of meaning that it should inform
%inference at all levels of granularity. 

There are few hybrid systems that integrate logical and distributional
information, and we discuss some of them below. 

\namecite{beltagy:starsem13} transform distributional similarity to
weighted distributional inference rules that are combined with
logic-based sentence representations, and use probabilistic
inference over both. This is the approach that we build on in this paper. 
\namecite{lewis:tacl13}, on the other hand, use clustering on
distributional data to infer word senses, and perform standard first-order
inference on the resulting logical forms. The main difference between
the two approaches lies in the role of gradience. Lewis and Steedman
view weights and probabilities as a problem to be avoided. We believe
that the uncertainty inherent in both language processing and world
knowledge should be front and center in all inferential processes. 
\namecite{Tian:2014vh} represent sentences using Dependency-based
Compositional
Semantics~\cite{liang-jordan-klein:2011:ACL-HLT2011}. They construct
phrasal entailment rules based on a logic-based alignment, and use
distributional similarity of aligned words to filter rules that do not
surpass a given threshold. 

Also related are distributional models where the dimensions of the
vectors encode model-theoretic structures rather than observed
co-occurrences~\cite{Clark:2012un,Sadrzadeh:2013kw,grefenstette:starsem13,herbelot:emnlp15},
even though they are not strictly hybrid systems as they do not
include contextual distributional information. \namecite{grefenstette:starsem13} 
 represents logical constructs using vectors and tensors, 
 but concludes that they do not adequately capture logical structure,
 in particular quantifiers. 

If like \namecite{AndrewsViglioccoVinson:09},
\namecite{Silberer:2012ug} and \namecite{bruni-EtAl:2012:ACL2012}
(among others) we also consider \emph{perceptual} context as part
of distributional models, then \namecite{CooperLappin:LiLT} also qualifies
as a hybrid logical/distributional approach. They envision a
classifier that labels feature-based representations of situations
(which can be viewed as perceptual distributional representations) as
having a certain probability of making a proposition true, for example
$smile(Sandy)$. These propositions function as types of situations in
a type-theoretic semantics.

%\paragraph{Reasons for combining logic and distributional models}
%More importantly, however, the combination of logical form and
%distributional models allows us to study 
%individual phenomena in depth. In this
%paper, the questions that we address include the best representation
%of quantifiers when domains are fixed in size (Section~\ref{sec:dca}), the handling of negation
%in a textual entailment context (Section~\ref{sec:cwa}), and the
%question of how well compositional distributional representations work
%for phrasal entailment (Section~\ref{subsubsec:phrasal}). In the
%future, we also want to explore representing determiners like
%\textit{most} in a probabilistic framework, and we very much want to
%use distributional representations of word
%meaning in context. The RTE task does not present many
%opportunities to address polysemy, but we expect this issue
%to become more prominent as we move to question answering.

\begin{figure}[tb]
  \centering
\includegraphics[width=0.4\columnwidth]{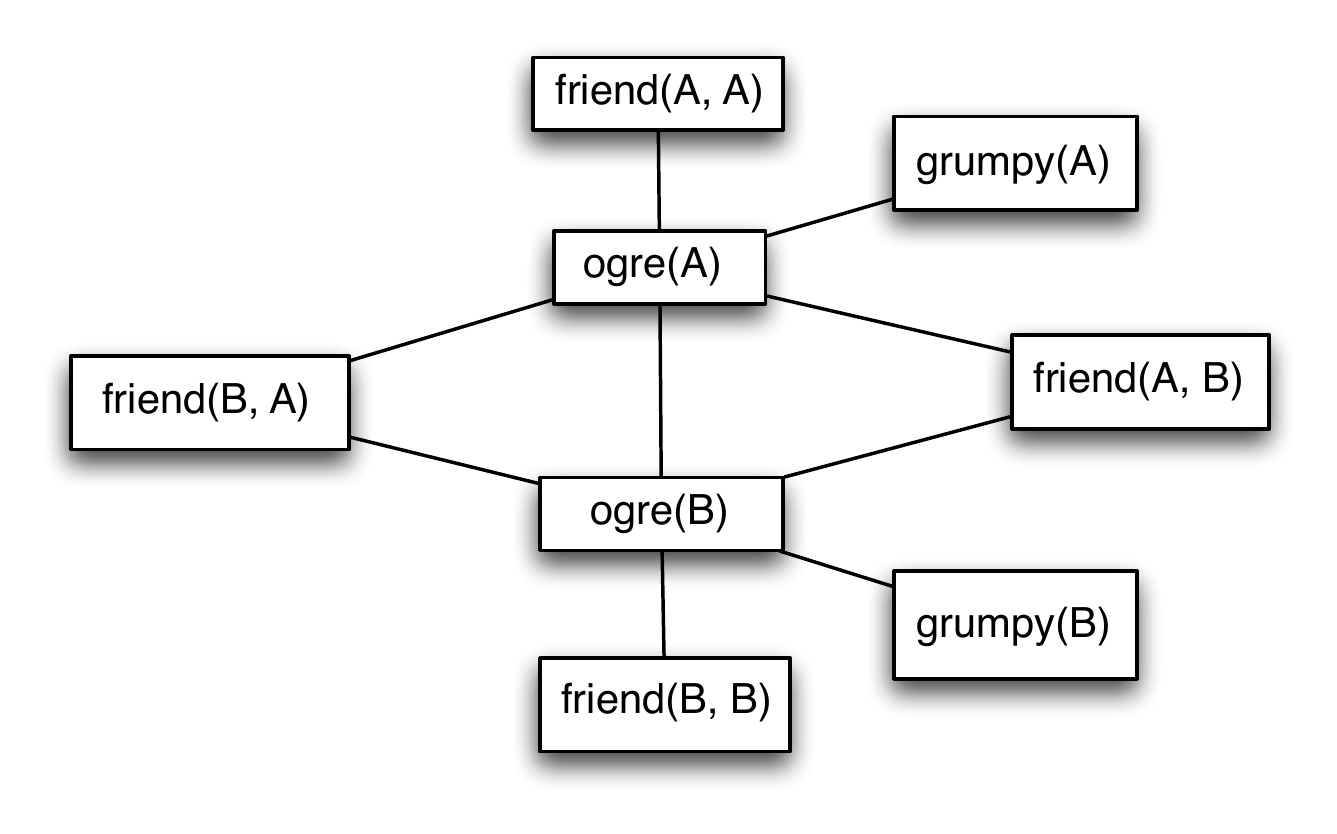}
\caption{A sample ground network for a Markov Logic Network}
 \label{fig:mln_ogre}
\end{figure}

\paragraph{Probabilistic Logic with Markov Logic Networks}
To combine logical and probabilistic information,  we
utilize Markov Logic Networks (MLNs) \cite{richardson:mlj06}.
MLNs are well suited for our
approach since they provide an elegant framework for assigning weights to
first-order logical rules, combining a diverse set of inference rules and
performing sound probabilistic inference.

A weighted rule allows truth assignments in which not all instances of
the rule hold. Equation~\ref{eq:smoke} above shows sample weighted
rules: Friends of ogres tend to be ogres and ogres tend
to be grumpy. Suppose we have two constants, Anna ($A$) and Bob
($B$). Using these two constants and the predicate symbols in
Equation~\ref{eq:smoke}, the  set of all ground atoms we can
construct is:
\begin{equation*}
\begin{aligned}
L_{A, B} = \{ & ogre(A), ogre(B), grumpy(A), grumpy(B), friend(A, A),\\ 
					  &friend(A, B), friend(B, A), friend(B, B)\}
 \end{aligned}
 \end{equation*}
If we only consider models over a domain with these two constants as entities,
then each truth assignment to $L_{A, B}$ corresponds to a model. 
MLNs make the assumption of a one-to-one correspondence
between constants in the system and entities in the domain. We 
discuss the effects of this \emph{domain closure assumption} below.

Markov Networks or undirected graphical models~\cite{pearl:book88} compute the
probability $P(X = x)$ of an assignment $x$ of values to the sequence
$X$ of all variables in the
model based on clique potentials, where a clique potential is a function that assigns a value to
each clique (maximally connected subgraph) in the graph. Markov Logic
Networks construct Markov Networks (hence their name) based on weighted first
order logic formulas, like the ones in Equation~\ref{eq:smoke}. Figure~\ref{fig:mln_ogre} shows the network for
Equation~\ref{eq:smoke} with two constants. Every ground atom
becomes a node in the graph, and two nodes are connected if they
co-occur in a grounding of an input formula. In this graph, each clique
corresponds to a grounding of a rule. For example, the clique
including $friend(A, B)$, $ogre(A)$, and $ogre(B)$ corresponds to the
ground rule $friend(A, B) \wedge ogre(A) \Rightarrow ogre(B)$. A
variable assignment $x$ in this graph assigns to each node a
value of either True or False, so it is a truth assignment (a world). The clique
potential for the clique involving  $friend(A, B)$, $ogre(A)$, and
$ogre(B)$ is $\exp(1.1)$ if $x$ makes the  ground rule true, and 0 otherwise.
This allows for nonzero probability for worlds $x$ in which not all
friends of ogres are also ogres, but it assigns exponentially more
probability to a world for each ground rule that it satisfies. 
% With the weighted rules, a set of constants need to be specified. For the rules in equation 
% \ref{eq:smoke}, we can add constants representing two persons, Anna ($A$) and Bob ($B$). 
% Probabilistic logic uses the constants to ``ground'' atoms with variables, so we get 
% ``ground atoms'' like $Smoke(A)$, $Smoke(B)$, $Cancer(A)$, $Cancer(B)$, 
% $Friend(A, A)$, $Friend(A, B)$, $Friend(B,A)$, $Friend(B, B)$. Rules are also 
% grounded by replacing each atom with variables by all its possible ground atoms. 

More generally, an MLN takes as input a set of weighted
first-order formulas $F = F_1, \ldots, F_n$ and a set $C$ of
constants, and constructs an
undirected graphical model in which the set of nodes is the set of
ground atoms constructed from $F$ and $C$.
It computes the probability distribution
$P(X=x)$ over worlds based on this undirected graphical
model. The probability of a
world (a truth assignment) $x$ is defined as:
\begin{equation}
\label{eq:mln}
P(X=x) = \dfrac{1}{Z} \exp \left( \sum_{i}{w_i n_i \left( x \right)} \right)
\end{equation}
where $i$ ranges over all formulas $F_i$ in $F$, $w_i$ is the weight
of $F_i$, $n_i(x)$ is the number of groundings of $F_i$ that are true
in the world $x$, and $Z$ is the partition function (i.e., it
normalizes the values to probabilities).
So the probability of a world increases
exponentially with the total weight of the ground clauses that it
satisfies.

Below, we use $R$ (for rules) to denote the input set of weighted
formulas. In addition, an MLN takes as input an evidence set $E$
asserting truth values for some ground clauses. For example, $ogre(A)$ means that Anna is an
ogre. Marginal inference for MLNs calculates the probability 
$P(Q|E,R)$ for a query formula $Q$. 

Alchemy~\cite{kok:tr05} is the most widely used MLN implementation. It is a
software package that contains implementations of a variety of MLN
inference and learning algorithms. However, developing a scalable, general-purpose, accurate
inference method for complex MLNs is an open problem.
MLNs have been used for various NLP applications including unsupervised coreference resolution~\cite{poon:emnlp08},
semantic role labeling~\cite{riedel:conll08} and event extraction~\cite{riedel:bionlp09}.
 
\paragraph{Recognizing Textual Entailment}
The task that we focus on in this paper is Recognizing Textual
Entailment (RTE)~\cite{dagan:hlt13},  the task of 
determining whether one natural language text, the \textit{Text  $T$}, \emph{entails}, 
\emph{contradicts}, or is not related (\emph{neutral}) to another, the \textit{Hypothesis $H$}. 
``Entailment''
here does not mean logical entailment: The Hypothesis is
entailed if a human annotator judges that it plausibly follows from
the Text. When using naturally occurring sentences, this is a very
challenging task that should be able
to utilize the unique strengths of both logic-based and distributional
semantics. 
Here are examples from the SICK dataset~\cite{marelli:lrec14}: 
\begin{itemize} [noitemsep]
\item Entailment
\item [T:] A man and a woman are walking together through the woods.
\item [H:] A man and a woman are walking through a wooded area.
\end{itemize}
\begin{itemize} [noitemsep]
\item Contradiction
\item [T:] Nobody is playing the guitar
\item [H:] A man is playing the guitar
\end{itemize}
\begin{itemize} [noitemsep]
\item Neutral
\item [T:] A young girl is dancing
\item [H:] A young girl is standing on one leg
\end{itemize}

The SICK (``Sentences Involving Compositional Knowledge'')
dataset, which we use for evaluation in this paper,  was designed to foreground particular linguistic
phenomena but to eliminate the need for world knowledge beyond
linguistic knowledge. It was constructed from sentences from two image
description datasets,
ImageFlickr\footnote{\url{http://nlp.cs.illinois.edu/
    HockenmaierGroup/data.html}} and the SemEval 2012 STS MSR-Video
Description data.\footnote{\url{http://www.cs.york.ac.uk/semeval-2012/
    task6/index.php?id=data}} Randomly selected sentences from these
two sources were first simplified to remove some linguistic phenomena
that the dataset was not aiming to cover. Then additional sentences
were created as variations over these sentences, by
paraphrasing, negation, and reordering. RTE pairs were then created
that consisted of a simplified original sentence paired with one of
the transformed sentences (generated from either the same or a
different original sentence). 

We would like to mention two particular systems
that were evaluated on SICK.  
The first is \namecite{lai:semeval14} which was the top performing system at the original shared task. They use
a linear classifier with many hand crafted features, including alignments, word forms,
POS tags, distributional similarity, WordNet, and a unique feature called Denotational Similarity. Many of
these hand crafted features are later incorporated in our lexical entailment classifier, described in
section \ref{sec:ent}. The Denotational Similarity uses a large database of human- and
machine-generated image captions to cleverly capture some world knowledge of entailments.

The second system is \namecite{bjerva:2014semeval} which  also participated in the original SICK shared task, and 
achieved 81.6\% accuracy. The RTE system uses Boxer to parse input sentences to logical form, then uses a theorem 
prover and a model builder to check for entailment and contradiction. The knowledge bases used are WordNet and PPDB.
In contrast with our work, PPDB paraphrases are not translated to logical rules (section \ref{sec:ppdb}). Instead, 
in case a PPDB paraphrase rule applies to a pair of sentences, the rule is applied at the text level before parsing the sentence. 
Theorem provers and model builders have high precision detecting entailments and contradictions, but low recall.
To improve recall, neutral pairs are reclassified using a set of textual, syntactic and semantic features.

\section{System Overview}
\label{sec:overview}

\begin{figure}[tb]
  \centering
\includegraphics[width=\textwidth]{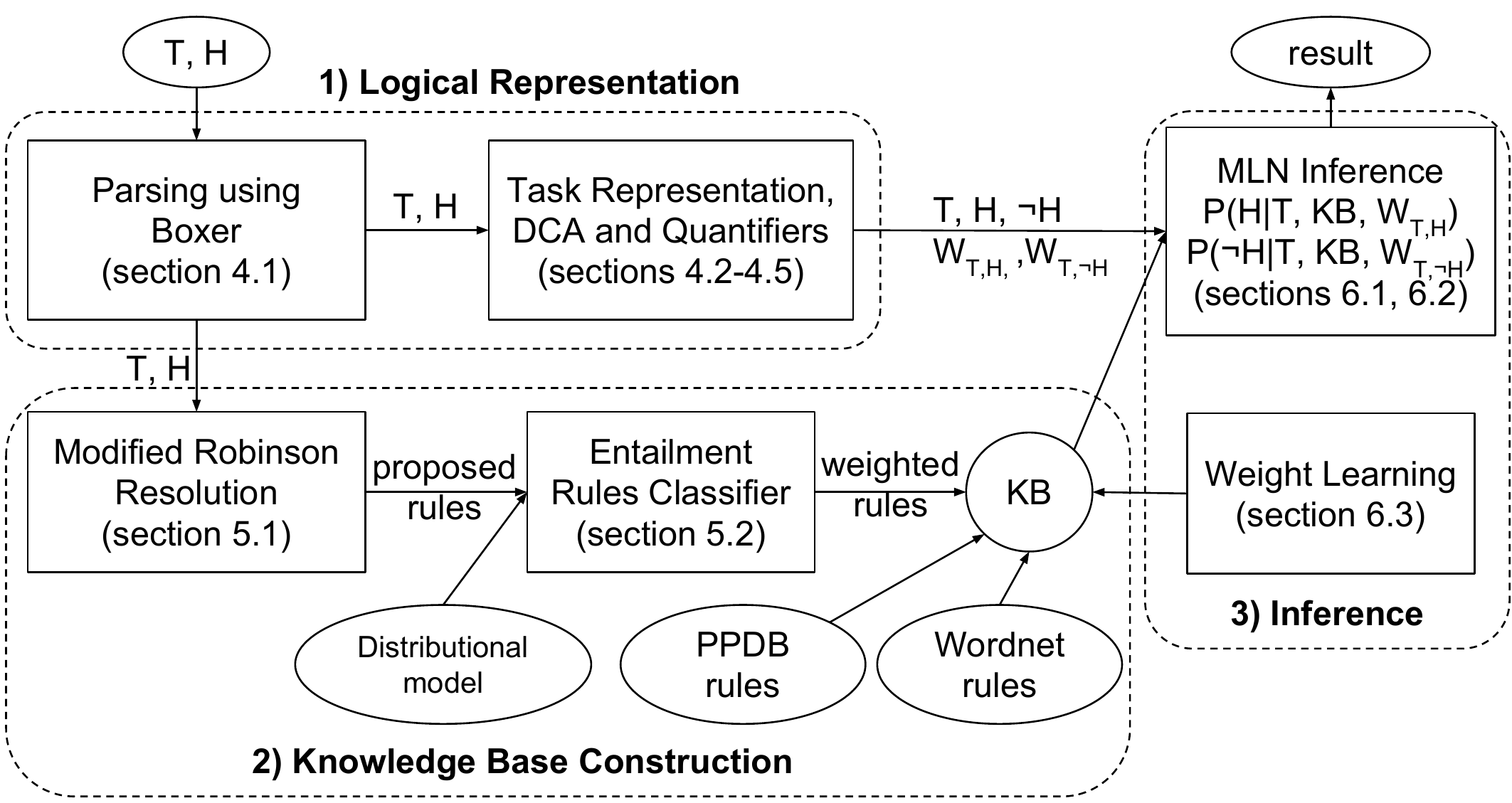}
\caption{System Architecture}
 \label{fig:arch}
\end{figure}

This section provides an overview of our system's architecture, using the following  RTE example to demonstrate 
the role of each component:
\begin{itemize}
\item [$T$:] A grumpy ogre is not smiling.
\item [$H$:] A monster with a bad temper is not laughing. 
\end{itemize}
Which in logic are: 
\begin{itemize}
\item [$T$:] $\exists x. \: ogre(x) \wedge grumpy(x) \wedge \lnot \exists y. \: agent(y,x) \wedge smile(y) $
\item [$H$:] $\exists x,y. \: monster(x) \wedge with(x,y) \wedge bad(y) \wedge temper(y) \wedge \lnot \exists z. \: agent(z,x) \wedge laugh(z)$. 
\end{itemize}
This example needs the following rules in the knowledge base $KB$:
\begin{itemize} 
\item [$r_1$:] laugh $\Rightarrow$ smile
\item [$r_2$:] ogre $\Rightarrow$ monster
\item [$r_3$:] grumpy $\Rightarrow$ with a bad temper
\end{itemize}

Figure \ref{fig:arch} shows the high-level architecture of our system and Figure \ref{fig:mlns} shows the MLNs
constructed by our system for the given RTE example.

\begin{figure}[h]
\begin{subfigure}{\textwidth}
 \begin{equation*}
 \openup-0.6\jot
\begin{aligned}
D: \: &\{O, L, C_o\}\\
G: \: &\{ogre(O), grumpy(O), monster(O), agent(L,O), smile(L), laugh(L), \\
&skolem_f(O, C_o), with(O, C_o), bad(C_o), temper(C_o) \}\\
T: \: &ogre(O) \wedge grumpy(O) \wedge \lnot \exists y. \: agent(y,O) \wedge smile(y)  \: | \:  \infty \\
r_1: \: &\forall x. \: laugh(x) \Rightarrow smile(x) \: | \:  w_1 \times w_{ppdb} \\
r_2: \: &\forall x. \: ogre(x) \Rightarrow monster(x) \: | \:   w_{wn} = \infty \\
r_3: \: &\forall x. \: grumpy(x) \Rightarrow  \forall y. \:  skolem_f(x,y) \Rightarrow with(x,y) \wedge bad(y) \wedge temper(y) \: | \\ & \:  w_3 \times w_{eclassif} \\
sk: \: &skolem_f (O, C_o) \: | \:  \infty  \\
A: \: &\forall x. \: agent(L,x) \wedge laugh(L) \: | \:  1.5 \\
H: \: &\exists x,y. \: monster(x) \wedge with(x,y) \wedge bad(y) \wedge temper(y) \wedge \lnot \exists z. \: agent(z,x) \wedge laugh(z)
 \end{aligned}
 \end{equation*}
 \caption{MLN to calculate $P(H|T,KB,W_{T, H})$ }
 \label{fig:pht}
\end{subfigure}%

\begin{subfigure}{\textwidth}
 \begin{equation*}
 \openup-0.6\jot
\begin{aligned}
D: \: &\{O, C_o, M, T\}\\
G: \: &\{ogre(O), grumpy(O), monster(O), skolem_f(O, C_o), with(O, C_o), bad(C_o), \\
&temper(C_o), monster(M), with(M,T), bad(T), temper(T) \}\\
T: \: &ogre(O) \wedge grumpy(O) \wedge \lnot \exists y. \: agent(y,O) \wedge smile(y)  \: | \:  \infty \\
r_1: \: &\forall x. \: laugh(x) \Rightarrow smile(x) \: | \:  w_1 \times w_{ppdb} \\
r_2: \: &\forall x. \: ogre(x) \Rightarrow monster(x) \: | \:   w_{wn} = \infty \\
r_3: \: &\forall x. \: grumpy(x) \Rightarrow  \forall y. \:  skolem_f(x,y) \Rightarrow with(x,y) \wedge bad(y) \wedge temper(y) \: | \\	& \:  w_3 \times w_{eclassif} \\
sk: \: &skolem_f (O, C_o) \: | \:  \infty  \\
A: \: &monster(M) \wedge with(M,T) \wedge bad(T) \wedge temper(T) \: | \:  1.5 \\
\lnot H: \: &\lnot \exists x,y. \: monster(x) \wedge with(x,y) \wedge bad(y) \wedge temper(y) \wedge \lnot \exists z. \: agent(z,x) \wedge laugh(z)\\
 \end{aligned}
 \end{equation*}
\caption{MLN to calculate $P(\lnot H|T,KB,W_{T, \lnot H})$ }
\label{fig:pnotht}
\end{subfigure}
\caption{MLNs for the given RTE example. 
The RTE task is represented as two inferences $P(H|T,KB,W_{T,H})$ 
and $P(\lnot H|T,KB,W_{T,\lnot H})$ (Section~\ref{sec:hGivenT}).
$D$ is the set of constants in the domain.
$T$ and $r_3$  are skolemized and $sk$ is the skolem function of $r_3$ (Section \ref{sec:dca}).
$G$ is the set of non-False (True or unknown) ground atoms as determined by the CWA (Section \ref{sec:cwa}, \ref{sec:mcw}).
$A$ is the CWA for the negated part of $H$ (Section \ref{sec:anticwa}).
$D, G, A$ are the world assumptions $W_{T,H}$ ( or $W_{T,\lnot H}$). 
$r_1, r_2, r_3$ are the $KB$.
$r_1$ and its weight $w_1$ are from PPDB (Section \ref{sec:ppdb}).
$r_2$ is from WordNet (Section \ref{sec:wn}).
$r_3$ is constructed using the Modified Robinson Resolution (Section \ref{sec:rr}), 
and its weight $w_3$ is calculated using the entailment rules classifier (Section \ref{sec:ent}).
The resource specific weights $w_{ppdb}, w_{eclassif}$ are learned using weight learning (Section \ref{sec:wlearn}).
Finally the two probabilities are calculated using MLN inference where $H$ (or $\lnot H$) 
is the query formula (Section \ref{sec:qf})
}
\label{fig:mlns}
\end{figure}

Our system has three main components:
\begin{enumerate}
  \item Logical Representation (Section \ref{sec:parse}), where input natural sentences $T$ and $H$ are
    mapped into logic then used to represent the RTE task as a probabilistic
    inference problem.

  \item Knowledge Base Construction $KB$ (Section \ref{sec:kb}), where the background knowledge is collected
  from different sources, encoded as first-order logic rules, weighted and added to the inference problem. 
  This is  where distributional information is integrated into our system.
  
  \item Inference (Section \ref{sec:infer}), which uses MLNs to solve
    the resulting inference problem.
\end{enumerate}

One powerful advantage of using a general-purpose probabilistic logic as a
semantic representation is that it allows for a highly modular system. Therefore,
the most recent advancements in any of the system components, in
parsing, in knowledge base resources and distributional semantics, and in
inference algorithms, can be easily incorporated into the system.

In the Logical Representation  step (Section \ref{sec:parse}), we map input sentences $T$ and $H$ to
logic. Then, we show how to map the three-way RTE classification (entailing, neutral,
or contradicting) to probabilistic inference problems.
The mapping of sentences to logic differs from standard
first order logic in several respects because of properties of the
probabilistic inference system. First, MLNs make the
Domain Closure Assumption (DCA), which states that there are no objects in the
universe other than the named constants~\cite{richardson:mlj06}. This
means that constants need to be explicitly introduced in the domain in
order to make probabilistic logic produce the expected
inferences. Another representational issue that we discuss is why we should make 
the closed-world assumption, and its implications on the task representation.

In the Knowledge Base Construction step $KB$ (Section \ref{sec:kb}), we collect inference rules from a variety
of sources. We add rules from existing databases, in particular 
WordNet~\cite{princeton:url10} and PPDB~\cite{ganitkevitch:naacl13}. To
integrate distributional semantics, we use a variant of Robinson
resolution to align the Text $T$ and the Hypothesis $H$, and to find
the difference between them, which we formulate as an entailment rule. We then train a lexical and phrasal
entailment classifier to assess this rule. 
Ideally, rules need be contextualized to handle polysemy, but we leave that to future work.

%Note that ideally, rules need be contextualized to handle
%polysemy and homonymy but we leave that to future work.

In the Inference step (Section \ref{sec:infer}), automated reasoning for MLNs is used to perform the RTE
task.  We implement an MLN inference algorithm that directly supports querying
complex logical formula, which is not supported in the available MLN
tools~\cite{beltagy:starai14}. We exploit the closed-world assumption
to help reduce the size of the inference problem in order to make it
tractable~\cite{beltagy:starai14}.  We also discuss weight learning for the
rules in the knowledge base.

\section{Logical Representation}
\label{sec:parse}
The first component of our system parses sentences into logical form and uses this
to represent the RTE problem as MLN inference.
We start with Boxer~\cite{bos:step08},  a rule-based semantic analysis system that
translates a CCG parse into a logical form.
The formula
\begin{equation}
\exists x, y, z. \: ogre(x) \wedge agent(y, x) \wedge love(y) \wedge patient(y, z) \wedge princess(z)
\label{eq:boxer2}
\end{equation}
is an example of Boxer producing discourse representation structures
using a neo-Davidsonian framework. 
We call Boxer's output alone an ``uninterpreted
logical form'' because the predicate symbols are simply words and do not have
meaning by themselves.  Their semantics derives from the knowledge base $KB$ we
build in Section~\ref{sec:kb}.
The rest of this section discusses how we adapt Boxer output for MLN inference.

\subsection{Representing Tasks as Text and Query}
\label{sec:hGivenT}

\paragraph{Representing Natural Language Understanding Tasks} 
In our framework, 
a language understanding task consists of a \emph{text} and a
\emph{query}, along with a \emph{knowledge base}. 
The text describes some situation or setting, and the query in the
simplest case asks whether a particular statement is true of the
situation described in the text. 
%\textbf{evaluates a particular statement given the situation described in the text, 
%where evaluation is a task dependent procedure. }
%asks whether a particular statement is true of the situation described in the text.
The knowledge base encodes relevant background knowledge: lexical
knowledge, world knowledge, or both.
In the textual entailment task, 
the text is the Text $T$, and the query is the
Hypothesis $H$.
%: In any situation described by $T$, would a human annotator view $H$ as being true? 
The sentence similarity (STS) task can be described as two
text/query pairs. In the first pair, the first sentence is the text and the
second is the query, and in the second pair the roles are reversed~\cite{beltagy:acl14}.
In question answering, the input documents constitute the text and
the query has the form $H(x)$
for a variable $x$; and the answer is the entity $e$ such that $H(e)$
has the highest probability given the information in $T$. 
%and the question 
%\textbf{is the query, and the answer is the entity that best fits the query.  }
%can be turned into query.
%%DELETED%%
%In all cases, distributional information 
%contributes to the knowledge base. In this paper, we
%will use $T$ for the text and $H$ for the query,  to
%match the textual entailment task on which we
%focus. Section~\ref{sec:kb} discusses the 
%particular knowledge base that we use.

In this paper, we focus on the simplest form  of text/query inference,
which applies to both RTE and STS:
Given a text $T$ and query $H$, does the text entail the query given
the knowledge base $KB$?
In standard logic, we determine
entailment by checking whether $T \wedge KB \Rightarrow
H$. (Unless we need to make the distinction explicitly, we
overload notation and use the symbol $T$ for the logical form computed for
the text,  and $H$ for the logical form computed for the query.) The
probabilistic version is to calculate the
probability $P(H|T,KB,W_{T, H})$, where $W_{T, H}$ is a world configuration, 
which includes the size of the domain.
%-- as mentioned
%above, this needs to be fixed in MLNs. 
We discuss $W_{T, H}$ in
Sections~\ref{sec:dca} and \ref{sec:cwa}. While we focus on the
simplest form of text/query inference, more complex tasks such as
question answering still
have the probability  $P(H|T,KB,W_{T, H})$ as part of their calculations.
%When a task asks for an
%output that is not a probability but a categorial decision, as is the
%case for example for RTE, the most straightforward solution is to
%learn thresholds on the probability.

\paragraph{Representing Textual Entailment} 
%%DELETED%% for space 
%Up to this point, the specification we have given applies to any
%natural language understanding task. We now look at the task
%formulation specifically for RTE. Markov Logic networks accept
%weighted first-order formulas, but a weight can also be infinite to
%indicate a hard rule. Basically, we use the contents of the text $T$
%as hard knowledge, and add soft lexical rules from the knowledge
%base (with some qualifications discussed below). 

RTE asks for a categorical decision
between three categories, Entailment, Contradiction, and Neutral. A
decision about Entailment can be made by learning a threshold on the
probability $P(H|T,KB,W_{T, H})$. To differentiate between
Contradiction and Neutral, we additionally calculate the probability $P(\lnot H|T,KB,W_{T, \lnot H})$. If $P(H|T,KB,W_{T, H})$ is high
while $P(\lnot H|T,KB,W_{T, \lnot H})$ is low, this indicates entailment. 
The opposite case indicates contradiction. If the two probabilities values
are close, this means $T$ does not significantly affect the probability of $H$,
indicating a neutral case. To learn the thresholds for these decisions,
we train an SVM classifier with LibSVM's default parameters~\cite{cjlin:manual01} 
to map the two probabilities to the final decision. The learned mapping is always 
simple and reflects the intuition described above.

\subsection{Using a Fixed Domain Size}
\label{sec:dca}

MLNs compute a probability distribution over possible worlds,
as described in Section~\ref{sec:relwork}. When we describe a task
as a text $T$ and a query $H$, the worlds over which the MLN computes a
probability distribution are ``mini-worlds'', just large enough to
describe the situation or setting given by $T$. The probability
$P(H|T, KB, W_{T, H})$ then describes the probability that $H$ would
hold given the probability distribution over the worlds that possibly describe $T$.
\footnote{\namecite{CooperLappin:LiLT} criticize
  probabilistic inference frameworks based on a probability
  distribution over worlds as not feasible. But what they mean by a
  world is a maximally consistent set of propositions. So because
we use MLNs only to handle ``mini-worlds'' describing individual
situations or settings, this criticism does not apply to our
approach.} 
The use of
``mini-worlds''  is by necessity, as
MLNs can only handle worlds with a fixed domain size, where ``domain size'' is the number of constants in the domain.
(In fact, this same restriction holds for all current practical probabilistic
inference methods, including PSL~\cite{bach:uai13}.) 
%As sketched in
%Section~\ref{sec:relwork}, MLNs reason over a given set of weighted
%formulas by first computing all groundings of these formulas and then
%consider the probability of different truth assignments to the ground
%atoms. This set of ground atoms naturally depends on the set of
%constants used to compute the groundings. 

Formally, the influence of
the set of constants on the worlds considered by an MLN can be
described by the Domain Closure Assumption
(DCA, \cite{GeneserethNilsson:87,richardson:mlj06}): The only models considered
for a set $F$ of formulas are those for which the following three conditions
hold: (a) Different constants refer to different objects in the domain, (b) the
only objects in the domain are those that can be represented using the constant
and function symbols in $F$, and (c) for each function $f$ appearing in $F$,
the value of $f$ applied to every possible tuple of arguments is known, and is
a constant appearing in $F$. Together, these three conditions entail
that \emph{there is a one-to-one relation between objects in the domain and the
named constants of $F$}. When the set of all constants is known, it can be used
to ground predicates to generate the set of all ground atoms, which then
become the nodes in the graphical model. Different constant sets result in
different graphical models.  If no constants are explicitly introduced, the
graphical model is empty (no random variables).  

This means that to obtain an adequate representation of an inference
problem consisting of a text $T$ and query $H$, we need to introduce a
sufficient number of constants explicitly into the formula: The worlds
that the MLN considers need to have enough constants
to faithfully represent the situation in $T$ and not give the wrong
entailment for the query $H$. In what follows, we explain how we
determine an appropriate set of constants for the logical-form representations
of $T$ and $H$. The domain size that we determine is one of the two
components of the parameter $W_{T, H}$.
%\footnote{Although MLNs supports multiple entity types, we do not utilize it and 
%all the entities we introduce have the same type.}

\paragraph{Skolemization} We introduce some of the necessary constants through the
well-known technique of \emph{Skolemization}~\citep{Skolem}. It transforms a formula $\forall x_1\ldots x_n
\exists y. F$ to $\forall x_1\ldots x_n. F^*$, where
$F^*$ is formed from $F$ by replacing all free occurrences of $y$ in $F$ by a
term $f(x_1, \ldots, x_n)$ for a new function symbol $f$. If $n=0$, f
is called a \emph{Skolem constant}, otherwise a \emph{Skolem
  function}. 
Although Skolemization is a widely used technique in first-order logic, 
it is not frequently employed in probabilistic logic since many
applications do not require existential quantifiers. 

We use Skolemization on the text $T$ (but not the query $H$, as
we cannot assume a priori that it is true). 
For example, the logical expression in Equation
 \ref{eq:boxer2}, which represents the sentence \textit{T: An ogre loves a princess},  will be Skolemized to:
 \begin{equation}
 ogre(O) \wedge agent(L, O) \wedge love(L) \wedge patient(L, N) \wedge princess(N)
 \end{equation}
 where $O, L, N$ are Skolem constants introduced into the domain.

Standard Skolemization transforms existential quantifiers embedded under universal quantifiers to
Skolem functions. For example, for the text \textit{T: All ogres snore} and its
logical form $\forall x. \: ogre(x) \Rightarrow \exists y. \: agent(y,
x) \wedge snore(y)$ the standard Skolemization is $\forall x. \:
ogre(x) \Rightarrow agent(f(x), x)
\wedge snore(f(x))$. Per condition (c) of
the DCA above, if a Skolem function appeared in a formula, we would have to
 know its value for any constant in the domain, and this value would have to be
 another constant. To achieve this, we introduce a new predicate
 $Skolem_f$ instead of each Skolem function $f$, and for every constant that is an \textit{ogre}, we add an
 extra constant that is a \textit{loving} event. The example above then
 becomes: 
\begin{equation*}
\begin{gathered}
T: \forall x. \: ogre(x) \Rightarrow  \forall y. \: Skolem_f(x,y) \Rightarrow 
agent(y, x) \wedge  snore(y)
 \end{gathered}
  \end{equation*}
If the domain contains a single \textit{ogre}  $O_1$, then we introduce a
new constant $C_1$ and an atom $Skolem_f(O_1, C_1)$ to state 
that
the Skolem function $f$ maps the constant $O_1$ to the constant $C_1$. 

\paragraph{Existence} But how would the domain contain an \textit{ogre} $O_1$
in the case of the text \textit{T: All ogres snore},
$\forall x.ogre(x) \Rightarrow \exists y. agent(y, x) \wedge snore(y)  $? Skolemization does not introduce any
variables for the universally quantified $x$. We still introduce a
constant $O_1$ that is an \textit{ogre}. This can be justified by
pragmatics since the sentence presupposes that there are, in fact,
\textit{ogres}~\citep{Strawson:50,Geurts:07}. 
%By using this existential presupposition, we avoid the problem of empty graphical models. 
We use the sentence's parse to identify the universal quantifier's 
restrictor and body, then introduce entities representing the restrictor of the quantifier~\citep{Beltagy:IWCS15}. 
The sentence \textit{T: All ogres snore} effectively changes to  
\textit{T: All ogres snore, and there is an ogre}. At this point,
Skolemization takes over to generate a constant that is an \textit{ogre}. 
Sentences like \textit{T: There are no ogres} is a special case: For
such sentences, we do not generate evidence of an \textit{ogre}. In this case, the
non-emptiness of the domain is not assumed because the sentence explicitly
negates it.

\paragraph{Universal quantifiers in the query} The most serious problem with the DCA is that it affects the behavior of 
universal quantifiers in the query. 
Suppose we know that \textit{T: Shrek is a green ogre}, represented
with Skolemization as $ogre(SH) \wedge green(SH)$. Then we can conclude
that \textit{H: All ogres are green}, because by the DCA we are only 
considering models with this single constant which we
know is both an \textit{ogre} and \textit{green}. To address this
problem, we again introduce new constants.

We want a query \textit{H: All ogres are green} to be judged
true iff there is evidence that all \textit{ogres} will be \textit{green}, no matter how
many \textit{ogres} there are in the domain. So $H$ should follow from
\textit{$T_2$: All ogres are green} but not from \textit{$T_1$:
  There is a green ogre}. Therefore we
introduce a new constant $D$ for the query and assert $ogre(D)$
to test if we can then conclude that $green(D)$. The new evidence $ogre(D)$ prevents the query
from being judged true given $T_1$. Given $T_2$, the new \textit{ogre} $D$ will be inferred to be \textit{green}, in which
case we take the query to be true. Again, with a query such as
\textit{H: There are no ogres}, we do not generate any evidence for
the existence of an \textit{ogre}. 

\subsection{Setting Prior Probabilities}
\label{sec:cwa}

Suppose we have an empty text $T$, and the query \textit{H: A is an
  ogre}, where $A$ is a constant in the system. Without any additional
information, the worlds in which $ogre(A)$ is true are going to be as
likely as the worlds in which the ground atom is false, so
$ogre(A)$ will have a probability of 0.5. So without any text
$T$, ground atoms have a prior probability in MLNs that is not zero. 
This prior probability depends mostly on the size of the set $F$ of input formulas.
%This prior
%probability depends on the number of worlds, hence on the size of the
%domain and the set of predicate symbols. 
%\textbf{Islam, correct like
%  this? not exactly. Prior probability of ground atoms mostly determined by 
%  the rules in the MLN more than the domain size. If the MLN has no rules at all, 
%  all ground atoms will get 0.5 prior probability regardless of the domain size }
 The prior probability of an individual ground atom can be
influenced by a weighted rule, for example $ogre(A) \mid -3$, with a
negative weight, sets a low prior probability on $A$ being an
ogre. This is the second group of parameters that we encode in $W_{T, H}$: 
weights on ground atoms to be used to set prior probabilities. 

Prior probabilities are problematic for our probabilistic encoding of natural language understanding
problems. As a reminder, we probabilistically test for entailment by
computing the probability of the query given the text, or more
precisely $P(H|T, KB, W_{T, H})$. However, how useful this conditional
probability is as an indication of entailment depends on the prior
probability of $H$, 
%However, this conditional probability is very sensitive to 
$P(H|KB,W_{T, H})$. For example, if $H$ has a high prior 
probability, then a high conditional probability $P(H|T,KB,W_{T, H})$ 
does not add much information because it is not clear if the probability is high 
because $T$ really entails $H$, or because of the high prior
probability of $H$. In practical terms, we would not want to say that
we can conclude
from \textit{T: All princesses snore} that \textit{H: There is an
  ogre} just because of a high prior probability for the existence of ogres.

%%DELETED% rewritten below without mentioning the ratio
%We discuss two suggestions on how to solve this problem and 
%make the probability $P(H|T,KB,W_{T, H})$ less sensitive to $P(H|KB,W_{T, H})$. 
%The first is to use the ratio between the conditional and the prior
%probability of the query, $\frac{P(H|T,KB,W_{T, H})}{P(H|KB,W_{T, H})}$, 
%with the intuition that the absolute value of $P(H|T,KB,W_{T, H})$
%does not really matter, but what matters is how much adding $T$ changes 
%the probability of $H$ positively (indicating entailment) or 
%negatively (indicating contradiction).
%The second option is to pick a particular $W_{T, H}$ such that the prior 
%probability of $H$ is approximately zero, $P(H|KB,W_{T, H}) \approx 0$, so that 
%we know that any increase in the conditional probability is 
%an effect of adding $T$. For the task of RTE, where we need to
%distinguish Entailment, Neutral, and Contradiction, this inference alone does not account for contradictions, 
%which is why an additional inference $P(\lnot H|T,KB,W_{T, \lnot H})$
%is needed.
%
%For the rest of this section, we argue why we believe the first option is not a good 
%fit for natural language understanding tasks formulated in terms of a
%text and query, while the second is a better fit. Then we show how to set 
%the world configurations $W_{T, H}$ such that $P(H|KB,W_{T, H}) \approx 0$ 
%by enforcing the closed-world assumption (CWA). This is the assumption that 
%all ground atoms have very low prior probability (or are false by default).

To solve this problem and  make the probability $P(H|T,KB,W_{T, H})$ less sensitive to $P(H|KB,W_{T, H})$, 
we pick a particular $W_{T, H}$ such that the prior 
probability of $H$ is approximately zero, $P(H|KB,W_{T, H}) \approx 0$, so that 
we know that any increase in the conditional probability is 
an effect of adding $T$. For the task of RTE, where we need to
distinguish Entailment, Neutral, and Contradiction, this inference alone does not account for contradictions, 
which is why an additional inference $P(\lnot H|T,KB,W_{T, \lnot H})$
is needed.

For the rest of this section, we show how to set 
the world configurations $W_{T, H}$ such that $P(H|KB,W_{T, H}) \approx 0$ 
by enforcing the closed-world assumption (CWA). This is the assumption that 
all ground atoms have very low prior probability (or are false by default).

\paragraph{Using the CWA to set the prior probability of the query
  to zero}
The closed-world assumption (CWA) is the assumption that 
everything is false unless stated otherwise. We translate it to our probabilistic
setting as saying that all ground atoms have very 
low prior probability. For most queries $H$, setting the world configuration
$W_{T, H}$ such that all ground atoms have low prior probability 
is enough to achieve that $P(H|KB,W_{T, H}) \approx 0$
(not for negated $H$s, and this case is discussed below).
For example, \textit{H: An ogre loves a princess}, in logic is: 
 \begin{equation*}
  \begin{gathered}
H: \exists x,y,z. \:  ogre(x) \wedge agent(y,x) \wedge love(y) \wedge patient(y,z) \wedge princess(z) 
 \end{gathered}
\end{equation*}
Having low prior probability on all ground atoms means that the prior 
probability of this existentially quantified $H$ is close to zero.

%\paragraph{Reasons to set the prior probability of H to zero}
%\paragraph{Reasons to make the CWA}
We believe that this setup is more appropriate for probabilistic
natural language entailment
for the following reasons. 
First, this aligns with our intuition of what it means for a query to
follow from a text: that $H$ should be entailed by $T$ not because of
general world knowledge. For example, if \textit{T: An ogre loves a princess}, and 
\textit{H: Texas is in the USA}, then although $H$ is true in the real world, 
$T$ does not entail $H$. 
%Even though $H$ is true in the real world,
%$T$ is not the answer to the query. 
Another example: \textit{T: An ogre loves a princess}, 
\textit{H: An ogre loves a green princess}, again, $T$ does not entail $H$
because there is no evidence that the \textit{princess} is \textit{green}, 
in other words, the ground atom $green(N)$ has very low prior probability. 
%%DELETED% for space
%As we said above, we construct the
%worlds over which the MLN reasons to encode the situation or setting
%described by $T$. Anything that is not 
%explicitly stated in $T$ should be assumed to be false by default. 
%In the RTE task, this is an explicit part of
%the task specification. In sentence similarity (STS), we would not
%want a known fact like \textit{Texas is in the USA} to be judged
%similar to every sentence. In question answering (QA), a text should
%only count as an answer to a query if it actually addresses the query.

The second reason is that with the CWA, the inference result is less sensitive to 
the domain size (number of constants in the domain).
In logical forms for typical natural language sentences, most variables in the query are existentially quantified. 
Without the CWA, the probability of 
an existentially quantified query increases as the domain size increases, regardless of the evidence. 
This makes sense in the MLN setting, because in larger domains 
the probability that something exists increases. However, this is not what 
we need for testing natural language queries, as the probability of the query should 
depend on $T$ and $KB$, not the domain size. With the CWA, what affects the probability
of $H$ is the non-zero evidence that $T$ provides and $KB$, 
regardless of the domain size. 

%%DELETED%% summerized below
%The third reason is computational efficiency. 
%As discussed in Section~\ref{sec:relwork}, Markov Logic Networks
%perform probabilistic inference by first computing all possible
%groundings of a given set of weighted formulas, and then using a
%Markov network to compute probabilities for truth assignments over
%this set of ground atoms. The grounding step can require significant amounts of
%memory. This is particularly striking for problems in natural language semantics because
%they usually have mostly existentially quantified variables, which,
%through Skolemization, results in a model with many constants. \namecite{beltagy:starai14} show how to utilize the CWA to 
%address this problem by reducing the number of ground atoms that the system generates. 
%They determine, based on $KB$, which ground atoms 
%are useful to model (probability does not equal prior probability)
%and which are not (probability remains at the prior probability). 
%They use an algorithm that tracks the propagation of evidence from $T$
%through $KB$, which finds (without running inference) the ground atoms 
%whose probabilities will remain at their prior probability. 
%These ground atoms can be assumed to be false and can be dropped from the 
%inference problem without significantly changing the computed probability of $H$.
%We discuss this algorithm below in Section \ref{sec:mcw}. 

The third reason is computational efficiency. 
As discussed in Section~\ref{sec:relwork}, Markov Logic Networks
first compute all possible
groundings of a given set of weighted formulas which can require significant amounts of
memory. This is particularly striking for problems in natural language semantics because of long formulas. \namecite{beltagy:starai14} show how to utilize the CWA to 
address this problem by reducing the number of ground atoms that the system generates. 
We discuss the details in Section \ref{sec:mcw}.

\paragraph{Setting the prior probability of negated $H$ to zero}
\label{sec:anticwa}
While using the CWA is enough to set $P(H|KB,W_{T, H}) \approx 0$
for most $H$s, it does not work for \emph{negated} $H$ (negation is part of $H$). 
Assuming that everything is false by default and that all ground atoms 
have very low prior probability (CWA) means 
that all negated queries $H$ are true by default. 
The result is that all negated $H$ are judged entailed regardless of $T$.
For example, \textit{T: An ogre loves a princess} would entail 
\textit{H: No ogre snores}. This
$H$ in logic is:
 \begin{equation*}
  \begin{gathered}
%H: \lnot \exists x,y. \: young(x) \wedge girl(x) \wedge agent(y,x) \wedge dance(y)
H: \forall x,y. \: ogre(x) \Rightarrow \lnot ( agent(y,x) \wedge snore(y))
 \end{gathered}
\end{equation*}
As both $x$ and $y$ are universally quantified variables in $H$, 
we generate evidence of an ogre
$ogre(O)$ as described in section \ref{sec:dca}. 
Because of the CWA, $O$ is assumed to be \textit{does not snore}, 
and $H$ ends up being true regardless of $T$. 

To set the prior probability of $H$ to $\approx 0$ and prevent it from 
being assumed true when $T$ is
just uninformative, we construct
a new rule $A$ that implements a kind of anti-CWA.
$A$ is formed as a conjunction of 
all the predicates that were not used to generate evidence before, 
and are \emph{negated} in $H$.
%This rule $R$ will get high positive weight but not an infinite weight. 
This rule $A$ gets a positive weight indicating that its ground atoms have 
high prior probability. As the rule $A$ together with the evidence
generated from $H$ states the opposite of the negated parts of $H$,
the prior probability of $H$ is low, and $H$ cannot become true unless 
%infinite-weight information in $T$ explicitly negates $R$. 
$T$ explicitly negates $A$. $T$ is translated into unweighted rule,
which are taken to have infinite weight, and which thus can overcome
the finite positive weight of $A$. 
Here is a Neutral RTE example, % adapted from the SICK dataset, 
\textit{T: An ogre loves a princess}, and \textit{H: No ogre snores}. 
Their representations are:
\begin{itemize}
\item [$T$:] $\exists x,y,z. \: ogre(x) \wedge agent(y,x) \wedge love(y) \wedge patient(y,z) \wedge princess(z)$
%\item [$H$:] $\lnot \exists x,y. \: young(x) \wedge girl(x) \wedge agent(y,x) \wedge dance(y)$
\item [$H$:] $ \forall x,y. \: ogre(x) \Rightarrow \lnot ( agent(y,x) \wedge snore(y))$
\item [$E$:] $ ogre(O) $
\item [$A$:] $ agent(S,O) \wedge snore(S) | w = 1.5 $
\end{itemize}
$E$ is the evidence generated for the universally quantified variables
in $H$, 
and $A$ is the weighted rule for the remaining negated 
%universally quantified 
predicates. 
The relation between $T$ and $H$ is Neutral, as $T$ does not entail $H$. 
This means, we want $P(H|T,KB,W_{T, H}) \approx 0$, 
but because of the CWA, $P(H|T,KB,W_{T, H}) \approx 1$.
Adding $A$ solves this problem and $P(H|T,A,KB,W_{T, H}) \approx 0$
because $H$ is not explicitly entailed by $T$. 

In case $H$ contains existentially quantified variables that occur in
negated predicates, they need to be universally quantified in $A$ 
for $H$ to have a low prior probability.
For example, \textit{H: There is an ogre that is not green}: 
 \begin{equation*}
  \begin{gathered}
H: \exists x. \: ogre(x) \wedge \lnot green(x)  \\
A: \forall x. \: green(x) | w = 1.5
 \end{gathered}
\end{equation*}
If one variable is universally quantified and the other 
is existentially quantified, we need to do
something more complex. 
Here is an example, \textit{H: An ogre does not snore}:
 \begin{equation*}
  \begin{gathered}
H: \exists x. \: ogre(x) \wedge \lnot ( \: \exists y. \: agent(y,x)  \wedge  snore(y) \: ) \\
A: \forall v. \: agent (S, v) \wedge  snore(S) | w = 1.5
 \end{gathered}
\end{equation*}

\paragraph{Notes about how inference proceeds with the rule $A$ added}
If $H$ is a negated formula that is entailed by $T$, 
then $T$ (which has infinite weight) will contradict $A$, allowing $H$
to be true. Any weighted inference rules in the knowledge base $KB$ will need weights high
enough to overcome $A$. So the weight of $A$ is taken into account when
computing inference rule weights. 

In addition, adding the rule $A$ introduces constants in the domain that are necessary 
for making the inference. For example, take
\textit{T: No monster snores}, and 
\textit{H: No ogre snores}, 
which in logic are: 
\begin{itemize}
\item [$T$:] $\lnot \exists x,y. \: monster(x) \wedge agent(y,x) \wedge snore(y)$
\item [$H$:] $\lnot \exists x,y. \: ogre(x) \wedge agent(y,x) \wedge snore(y)$
\item [$A$:] $ogre(O) \wedge agent(S,O) \wedge snore(S) |  w = 1.5 $
\item [$KB$:] $\forall x. \: ogre(x) \Rightarrow monster(x) $
\end{itemize}
Without the constants $O$ and $S$ added by the rule $A$, the domain would have been empty
and the inference output would have been wrong. The rule $A$ prevents this problem.
In addition, the introduced 
evidence in $A$ fit the idea of ``evidence propagation'' mentioned above, 
 (detailed in Section \ref{sec:mcw}). For entailing sentences that are negated, like
 in the example above, the evidence propagates from $H$ to $T$ 
 (not from $T$ to $H$ as in non-negated examples). In the example, 
 the rule $A$ introduces  an evidence for $ogre(O)$ that then 
 propagates from the LHS to the RHS of the $KB$ rule.

\subsection{Textual Entailment and Coreference}
\label{sec:coref}

The adaptations of logical form that we have discussed so far apply to
any natural language understanding problem that can be formulated as
text/query pairs. The adaptation that we discuss now is specific to
textual entailment. It concerns coreference between text and query. 

For example, if we have \textit{T: An ogre does not snore} and 
\textit{H: An ogre snores}, then strictly speaking $T$ and $H$ are not contradictory 
because it is possible that the two sentences are referring to different \textit{ogres}. 
Although the sentence uses \textit{\textbf{an} ogre} not \textit{\textbf{the} ogre}, 
the annotators make the assumption that the \textit{ogre}  in $H$ 
refers to the \textit{ogre}  in $T$. 
In the SICK textual entailment dataset, many of the pairs that
annotators have labeled as contradictions are
only contradictions if we assume that some expressions corefer across $T$ and
$H$. 

For the above examples, here are the logical formulas 
with coreference in the $updated\: \lnot H$: 
 \begin{equation*}
\begin{aligned}
T&:  \exists x. \: ogre(x) \wedge \lnot (\exists y. \: agent(y,x) \wedge snore(y) ) \\
Skolemized\: T&:  ogre(O) \wedge \lnot (\exists y. \: agent(y,O) \wedge snore(y) ) \\
H&:  \exists x,y. \: ogre(x) \wedge agent(y,x) \wedge snore(y) \\
\lnot H &: \lnot \exists x,y. \: ogre(x) \wedge agent(y,x) \wedge snore(y) \\
updated\: \lnot H&:  \lnot \exists y. \: ogre(O) \wedge agent(y,O) \wedge snore(y)
 \end{aligned}
 \end{equation*}
Notice how the constant $O$ representing the \textit{ogre} in $T$ is used in the
$updated\: \lnot H$ instead of the quantified variable $x$. 

We use a rule-based approach to determining coreference between $T$
and $H$, considering both coreference between entities and coreference
of events. Two items (entities or events) corefer
if they 1) have different polarities, and 2) share the same lemma or share an inference rule. 
Two items have different polarities in $T$ and $H$ if  one of
them is embedded under a negation and the other is not. 
%Among the pairs of items that fulfill this condition, the simplest
%case of coreference is if a pair of items has the same lemma. 
For the example above, \textit{ogre} in $T$ is not negated, and 
\textit{ogre} in $\lnot H$  is negated, and both words are the same, so they corefer. 
%Here is another example: In 
%\textit{T: An ogre loves a pretty princess} and 
%\textit{H: An ogre loves an ugly princess}, 
%all items in $\lnot H$ are negated, so the \textit{ogre}s, \textit{princess}es and 
%the \textit{loving} events are all coreferring. Of course we still need an inference rule 
%\textit{pretty $\Leftrightarrow \lnot$ ugly}, which we obtain from WordNet as explained in 
%Section~\ref{sec:wn}. 

A pair of items in $T$ and $H$ under different polarities can also
corefer if they share an inference rule. In the example of 
\textit{T: A monster does not snore} and 
\textit{H: An ogre snores}, we need \textit{monster} and \textit{ogre}
to corefer. For cases like this, we rely on the inference rules found using the
modified Robinson resolution method discussed in Section~\ref{sec:rr}.  In
this case, it determines that \textit{monster} and \textit{ogre} should be
aligned, so they are marked as coreferring.  Here is another example:
\textit{T: An ogre loves a princess}, 
\textit{H: An ogre hates a princess}. In this case, \textit{loves} and
\textit{hates} are marked as coreferring.

\subsection{Using multiple parses}
\label{sec:boxer}

In our framework that uses probabilistic inference followed by
a classifier that learns thresholds, we can easily incorporate
multiple parses to reduce errors due to misparsing. 
Parsing errors lead to errors in the logical form representation,
which in turn can lead to erroneous entailments. If we can obtain 
multiple parses for a text $T$ and query $H$, and hence multiple
logical forms, this should increase our chances of getting a good
estimate of the probability of $H$ given $T$. 

The default CCG parser that Boxer uses is C\&C
~\cite{clark:acl04}. This parser can be configured to produce multiple
ranked parses~\cite{ng:acl12}; however, we found that
the top parses we get from C\&C are usually not diverse enough and map to the
same logical form. Therefore, in addition to the top C\&C parse, we use the top
parse from another recent CCG parser, EasyCCG~\cite{lewis:emnlp14}. 

So for a natural language text $N_T$ and query $N_H$, we obtain two parses each, say $S_{T1}$
and $S_{T2}$ for $T$ and $S_{H1}$ and $S_{H2}$ for $H$, which are
transformed to logical forms $T_1, T_2, H_1, H_2$. We now compute
probabilities for all possible combinations of representations of
$N_T$ and $N_H$: the probability of $H_1$ given $T_1$, the probability
of $H_1$ given $T_2$, and conversely also the probabilities of $H_2$
given either $T_1$ or $T_2$. If the task is textual entailment with
three categories Entailment, Neutral, and Contradiction, then as
described in Section~\ref{sec:hGivenT} we also compute the probability
of $\lnot H_1$ given either $T_1$ or $T_2$, and the probability of
$\lnot H_12$ given either $T_1$ or $T_2$. When we use multiple parses
in this manner, the thresholding classifier is simply trained to take
in all of these probabilities as features. 
In
Section~\ref{sec:eval}, we evaluate using C\&C alone and using both parsers.

\section{Knowledge Base Construction}
\label{sec:kb}
This section discusses the automated construction of the knowledge base, which
includes the use of distributional information to predict lexical and phrasal
entailment. This section integrates two aims that are conflicting
to some extent, as alluded to in the introduction. The first is to
show that a general-purpose in-depth natural language understanding system
based on both logical form and distributional representations can be adapted to
perform the RTE task well enough to achieve state of the art results. To
achieve this aim, we build a classifier for lexical and phrasal entailment that
includes many task-specific features that have proven effective in
state-of-the-art
systems~\cite{marelli:semeval14,bjerva:2014semeval,lai:semeval14}.  The second
aim is to provide a framework in which we can test different distributional
approaches on the task of lexical and phrasal entailment as a building block in
a general textual entailment system. To achieve this second aim, in
Section~\ref{sec:eval})  we 
provide an in-depth ablation study and error analysis for the effect of
different types of distributional information within the lexical and phrasal
entailment classifier.

Since the biggest computational bottleneck for MLNs is the creation of the
network, we do not want to add a large number of inference rules blindly to a
given text/query pair. Instead, we first examine the text and query to determine
inference rules that are potentially useful for this particular entailment
problem. 
For pre-existing rule collections, we add all possibly matching
rules to the inference problem (Section~\ref{sec:precompiled}).
For more
flexible lexical and phrasal entailment, we use the text/query pair to
determine additionally useful inference rules, then automatically create and weight these
rules. We use a variant of Robinson resolution~\cite{robinson:jacm65} to
compute the list of useful rules (Section~\ref{sec:rr}), then apply a lexical
and phrasal entailment classifier (Section~\ref{sec:ent}) to weight them.

Ideally, the weights that we compute for inference rules should depend
on the context in which the words appear. After all, the ability to
take context into account in a flexible fashion is one of the biggest
advantages of distributional models. Unfortunately the textual
entailment data that we use in this paper does not lend itself
to contextualization -- polysemy just does not play a large role in
any of the existing RTE datasets that we have used so far. Therefore, we leave
this issue to future work.

\subsection{Robinson Resolution for Alignment and Rule Extraction}
\label{sec:rr}

To avoid undo complexity in the MLN, we only want to add inference rules
specific to a given text $T$ and query $H$. Earlier versions of our system generated
distributional rules matching any word or short phrase in $T$ with any word or
short phrase in $H$. This includes many unnecessary rules, for example for \textit{T: An ogre loves a princess} and
\textit{H: A monster likes a lady}, the system  generates
rules linking \textit{ogre} to \textit{lady}. In this paper, we use a novel method to generate only 
rules directly relevant to $T$ and $H$: We assume that $T$ entails $H$, and ask
what missing rule set $KB$ is necessary to prove this entailment. 
We use a variant of Robinson resolution~\cite{robinson:jacm65} to 
generate this $KB$. Another way of viewing this technique is 
that it generates an \emph{alignment} between words and phrases in $T$ and
words or phrases in $H$ guided by the logic. 

%% The method that we use here is applicable to the RTE and STS tasks, where
%% both the text and the query are single sentences and it makes sense to
%% compute an alignment. It is not applicable to QA, where the text can
%% consist of long passages. For QA, we will want to fall back on our previous strategy of listing
%% all rules that could possibly apply somewhere in the text or somewhere
%% in the query.

\paragraph{Modified Robinson Resolution}
Robinson resolution is a theorem proving method for testing unsatisfiability
that has been used in some previous RTE systems \namecite{Bos:09}.  It assumes
a formula in conjunctive normal form (CNF),
%a conjunction of disjunctions of literals. 
a conjunction of clauses, where a clause is a disjunction of literals, and 
a literal is a negated or non-negated atom.
More formally, the formula has the form 
$\forall x_1, \ldots, x_n\big(C_1 \wedge \ldots \wedge C_m)$, 
where  $C_j$ is a clause and it has the form $L_1 \vee \ldots \vee L_k  $
where $L_i$ is a literal, which is an atom $a_i$  or a negated atom $\lnot a_i$. 
The resolution rule takes two clauses containing
complementary literals, and produces a new clause implied by
them. Writing a clause $C$ as the set of its literals, we can
formulate the rule as:
\[
\frac{C_1 \cup \{L_1\} ~~~~~~~~ C_2 \cup \{L_2\}}{(C_1 \cup C_2)_{\theta}}
\]
where $\theta$ is a most general unifier of $L_1$ and $\neg L_2$. 

In our case, we use a variant of Robinson resolution to remove the
parts of text $T$ and query $H$ that the two sentences have in common. Instead
of one set of clauses, we use two: one is the CNF of $T$, the other is
the CNF of $\neg H$. The resolution rule is only applied to pairs of
clauses where one clause is from $T$, the other from $H$. 
When no further applications of the resolution rule are possible, we
are left with remainder formulas $rT$ and $rH$. If $rH$ contains the 
empty clause, then $H$ follows from $T$ without inference rules. Otherwise, inference rules need to be generated. In the simplest
case, we form a single inference rule as follows. All variables
occurring in $rT$ or $rH$ are existentially quantified, all constants
occurring in $rT$ or $rH$ are un-Skolemized to new universally
quantified variables, and we infer the negation of $rH$ from $rT$.
That is, we form the inference rule 
\[ \forall x_1 \ldots x_n \exists y_1 \ldots y_m. \: rT\overline{\theta}
\Rightarrow \lnot rH\overline{\theta}\]
where $\{y_1\ldots y_m\}$ is the set of all variables occurring in
$rT$ or $rH$,
$\{a_1, \ldots a_n\}$ is the set of all constants occurring in $rT$ or $rH$
and $\overline{\theta}$ is the inverse of a substitution $\theta: \{
a_1 \to x_1, \ldots, a_n \to x_n\}$ for distinct variables $x_1,
\ldots, x_n$. 

For example,
consider \textit{T: An ogre loves a princess} and \textit{H: A
  monster loves a princess}. This gives us the following two clause
sets. Note that all existential quantifiers have been eliminated
through Skolemization. The query is negated, so we get five
clauses for $T$ but only one for $H$. 
\[\begin{array}{rl}
T: & \{ogre(A)\}, \{princess(B)\}, \{love(C\}, \{agent(C, A)\},
    \{patient(C, B)\}\\
\neg H: & \{\neg monster(x), \neg princess(y), \neg love(z), \neg agent(z, x),
     \neg patient(z, y)\}
  \end{array}
  \]
The resolution rule can be applied 4 times. After that, $C$ has been
unified with $z$ (because we have resolved $love(C)$ with
$love(z)$), $B$ with $y$ (because we have resolved $princess(B)$ with
$princess(y)$), and $A$ with $x$ (because we have resolved $agent(C, A)$
with $agent(z, x)$). The formula $rT$ is $ogre(A)$, and $rH$ is
$\neg monster(A)$. So the rule that we generate is:
\[ \forall x. ogre(x) \Rightarrow monster(x)\]
The modified Robinson resolution thus does two things at once: It
removes words that $T$ and $H$ have in common, leaving the words for
which inference rules are needed, and it aligns words and phrases in
$T$ with words and phrases in $H$ through unification. 

%One important refinement to this general idea is that we need
%to distinguish meta-predicates introduced by Boxer, such as $agent(X,
%Y)$, from content predicates that correspond to words in the
%sentences. Resolving on meta-predicates can result in incorrect rules,
%for example in the case of \textit{T: A person solves a problem} and
One important refinement to this general idea is that we need
to distinguish content predicates that correspond to content words (nouns, verb and adjectives) in the
sentences from non-content predicates such as Boxer's meta-predicates $agent(X,Y)$.
Resolving on non-content predicates can result in incorrect rules,
for example in the case of \textit{T: A person solves a problem} and
\textit{H: A person finds a solution to a problem}, in CNF:
\[\begin{array}{rl}
T: & \{person(A)\}, \{solve(B)\}, \{problem(C)\}, \{agent(B, A)\},\{patient(B, C)\}\\
\neg H: & \{\neg person(x), \neg find(y), \neg solution (z), \neg problem(u), \neg agent(y, x), \neg patient(y,z),\\
   & \neg to(z, u)\}
  \end{array}
  \]
If we resolve $patient(B, C)$ with $patient(y, z)$, we
unify the problem $C$ with the solution $z$, leading to a wrong
alignment. We avoid this problem by resolving on non-content predicates %meta-predicates
only when they are fully grounded (that is, when the substitution of
variables with constants has already been done by some other
resolution step involving content predicates). 

In this variant of Robinson resolution, we currently do not perform
any search, but unify two literals {\it only} if they are fully grounded or
if the literal in $T$ has a \textit{unique} literal in $H$ that it can be
resolved with, and vice versa. This works for most pairs in the SICK dataset. 
In future work, we would like to add search
to our algorithm, which will help produce better rules for sentences with duplicate 
words. 

\paragraph{Rule Refinement}
The modified Robinson resolution algorithm gives us one rule per
text/query pair. This rule needs postprocessing, as it is sometimes
too short (omitting relevant context), and often it combines what
should be several inference rules. 

In many cases, a rule needs to be extended. This is the case when it only shows the difference between text and
query is too short and needs context to be usable as a
distributional rule, for example: 
\textit{T: A dog is running in the snow}, 
\textit{H: A dog is running through the snow}, the rule we get is
$\forall x, y. \: in(x, y) \Rightarrow through(x, y) $. 
Although this rule is correct, it does not carry enough information 
to compute a meaningful vector representation for each side. What we
would like instead is a rule that infers ``run through snow'' from
``run in snow''.

 Remember that the variables $x$ and $y$ were Skolem constants 
in $rT$ and $rH$, for example $rT: in(R, S)$ and $rH: through(R, S)$. 
We extend the rule by adding the content words that contain the constants $R$ and $S$. 
In this case, we add the \textit{running} event and the \textit{snow}
back in. 
The final rule is: 
$\forall x, y. \: run(x) \wedge in(x, y) \wedge snow(y) \Rightarrow run(x) \wedge through(x, y) \wedge snow(y) $. 
%%DELETED% for space
%Here is another example: 
%\textit{T: A person is pouring olive oil into a pot }, 
%\textit{H: A person is pouring cooking oil into a pot}, 
%and the rule is $\forall x. \: olive(x) \Rightarrow cooking(x) $
%which we extend to 
% $\forall x. \: olive(x) \wedge oil(x) \Rightarrow cooking(x) \wedge oil(x) $
 
%%DELETED% summerized below 
%In some cases however, extending the rule adds unnecessary complexity, for 
%example: 
%\textit{T: A man is jumping into an empty pool}, 
%\textit{H: A man is jumping into a full pool}
%and the rule is $\forall x. \: empty(x) \Rightarrow full(x)$, 
%and extending it gives 
%$\forall x. \: empty(x) \wedge pool(x) \Rightarrow full(x)\wedge pool(x)$ 
%which makes the rule unnecessary complex. 
%
%At the moment, we have no general algorithm for when to extend a rule, which would have to take 
%context into account. At this time, we
%extend all rules as described above. As discussed below, the
%entailment rules subsystem can itself choose to split long rules, and it may
%choose to split these extended rules again. 

In some cases however, extending the rule adds unnecessary complexity.  
However, we have no general algorithm for when to extend a rule, which would have to take 
context into account. At this time, we
extend all rules as described above. As discussed below, the
entailment rules subsystem can itself choose to split long rules, and it may
choose to split these extended rules again.

% In other examples, part of the rule better be extended and another part not, 
% for example: 
% \textit{T: A man is eating a bowl of cereal}, 
% \textit{H: A man is eating cereal}, 
% and the rule is: 
% $\forall x,y,z \: patient(x,y) \wedge bowl(y) \wedge  of(y,z) \Rightarrow patient(x,z)$, 
% which better be extended by adding the \textit{cereal} but not adding 
% the \textit{eating} event. 

Sometimes, long rules need to be split. A single pair $T$ and $H$ gives rise
to one single pair $rT$ and $rH$, which often conceptually represents
multiple inference rules. So we split $rT$ and $rH$ as follows. First, we split each formula into disconnected sets of predicates. 
For example, consider
\textit{T: The doctors are healing a man}, 
\textit{H: The doctor is helping the patient}
which leads to the rule 
$\forall x,y. \: heal(x) \wedge man(y) \Rightarrow help(x) \wedge patient(y)$. 
The formula $rT$ is split into $heal(x)$ and $man(y)$ because the two
literals do not have any variable in common and there is no relation
(such as $agent()$) to link them. Similarly, $rH$ is split into $help(x)$ and $patient(y)$. 
If any of the splits has more than one verb, we split it again, where each 
new split contains one verb and its arguments. 

After that, we create new rules that link any part of $rT$ to any part of
$rH$ with which it has at least one variable in common. So for our example we get  $\forall x \: heal(x) \Rightarrow help(x) $
and $\forall y \: man(y) \Rightarrow patient(y)$. 

% It is important to note that many rules could in principle be split even more, 
% for example a rule like
% \textit{a man and two women are facing a camera  $\Rightarrow$  a group of people are looking at the camera}  can be split into 
% \textit{a man and two women $\Rightarrow$  a group of people }  
% and 
% \textit{facing a camera  $\Rightarrow$  looking at the camera} . 
There are cases where splitting the rule 
does not work, for example with 
\textit{A person, who is riding a bike $\Rightarrow$ A biker }. 
Here, splitting the rule and using \textit{person $\Rightarrow$ biker}
loses crucial context information. So we do not perform those
additional splits at the level of the logical form, though the
entailment rules subsystem may choose to do further splits. 

%%DELETED%% briefly mentioned in the section below
%\subsubsection{Translating the logical rule into text}
%The output of our modified Robinson resolution is a logical formula. 
%We map this formula to a text before passing it to the 
%entailment rule classifier which currently expects textual rules.
%Each Boxer predicate or relation (except meta predicates and relations) 
%comes with an index pointing to the source word. 
%For each predicate or relation in the logical formula, we replace 
%it with its corresponding word from the original sentence. 
%This yields a simple readable rule text that 
%the entailment rules subsystem 
%can handle.

\paragraph{Rules as training data}
\label{sec:annotaterules}
The output from the previous steps is a set of rules $\{r_1, ..., r_n\}$ 
for each pair $T$ and $H$. One use of these rules is to test whether
$T$ probabilistically entails $H$. But there is a second use too: The
lexical and phrasal entailment classifier that we describe below is a
supervised classifier, which needs training data. So we use the
training part of the SICK dataset to create rules through modified
Robinson resolution, which we then use to train the lexical and
phrasal entailment classifier. For simplicity, 
we translate the Robinson resolution rules into textual  rules 
by replacing each Boxer predicate with its corresponding word. 

%% We need training data for three
%% possible labels on rules: entailing, non-entailing, and
%% contradictory. 

%% As an aside, we mentioned above that the modified
%% Robinson resolution strategy is applicable to the RTE and STS tasks,
%% but not QA. However, inference rule training data generated from an RTE task can be
%% used for any lexical and phrasal entailment classifier, no matter the
%% task to which it will be applied. And in fact we make the training
%% data that we derived from SICK freely available (see the introduction
%% for the link). 

Computing inference-rule training data from RTE
data requires deriving labels for
individual rules from the labels on RTE pairs
(Entailment, Contradiction and Neutral).
The Entailment cases are the most straightforward. Knowing that
 $T \wedge r_1 \wedge ... \wedge r_n \Rightarrow H$, 
 then it must be that all $r_i$ are entailing. 
 We automatically label all $r_i$ of the entailing pairs as entailing rules. 
 
 For Neutral pairs, we know that $T \wedge r_1 \wedge ... \wedge r_n \nRightarrow H$, 
 so at least one of the $r_i$ is non-entailing. We experimented with
 automatically labeling  all $r_i$ as non-entailing, but that adds a lot of
 noise to the training data. For example, if \textit{T: A man is
   eating an apple} and \textit{H: A guy is eating an orange}, then
 the rule \textit{man $\Rightarrow$ guy} is entailing, but the rule
 \textit{apple $\Rightarrow$ orange} is non-entailing.  So we
 automatically compare the
 $r_i$ from a Neutral pair to the entailing rules derived from entailing
 pairs. All rules $r_i$ found among the entailing rules from entailing
 pairs are assumed to be entailing (unless
 $n=1$, that is, unless we only have one rule), and all other rules
 are assumed to be non-entailing. We found that this step improved the accuracy 
of our system. To further improve the accuracy, we performed a manual
annotation of rules derived from Neutral pairs, focusing only on the
rules that do not appear in Entailing. We labeled rules as either
entailing or non-entailing. From around 5,900 unique rules, we found 
737 to be entailing. In future work, we plan to use multiple instance
learning~\cite{dietterich:ai97,Bunescu:2007ug} to avoid manual
annotation; we discuss this further in Section~\ref{sec:future}. 

For Contradicting pairs, we make a
few simplifying assumptions that fit almost  
all such pairs in the SICK dataset. In most of the contradiction
pairs in SICK, one of the two sentences $T$ or $H$ is negated. 
For pairs where $T$  or $H$ has a negation, we assume that this negation 
is negating the whole sentence, not just a part of it. 
We  first consider the case where $T$ is not negated, and $H = \lnot S_h$. 
As $T$ contradicts $H$, it must hold that $T \Rightarrow \lnot H$, 
so $T \Rightarrow \lnot \lnot S_h$, and hence $T \Rightarrow S_h$. 
This means that  we just need to run our modified Robinson resolution 
with the sentences $T$ and $S_h$ and label all resulting $r_i$ as entailing. 

Next we consider the case where $T = \lnot S_t$ while $H$ is not negated. 
As $T$ contradicts $H$, it must hold that $\lnot S_t \Rightarrow \lnot H$, 
so $H \Rightarrow S_t$. Again, this means that we just need to run 
the modified Robinson resolution 
with $H$ as the ``Text'' and $S_t$ as the ``Hypothesis'' and label all
resulting $r_i$ as entailing. %Because of quantifier polarity, the inferences
% here need to go from $H$ to $T$, not the other way around.

The last case of contradiction is when both $T$ and $H$ are not negated, 
for example: 
\textit{T: A man is jumping into an empty pool}, \textit{H: A man is jumping into a full pool}, where \textit{empty} and \textit{full} are antonyms. 
As before, we run the modified Robinson resolution with $T$ and $H$ 
and get the resulting $r_i$. 
Similar to the Neutral pairs, at least one of the $r_i$ is a
contradictory rule, while 
the rest could be entailing or contradictory rules. 
As for the Neutral pairs, we take a rule $r_i$ to be entailing if 
it is among the entailing rules derived so far. All
other rules are taken to be contradictory rules. We did not 
do the manual annotation for these rules because they are few.

\subsection{The Lexical and Phrasal Entailment Rule Classifier}
\label{sec:ent}

After extracting lexical and phrasal rules using our modified Robinson resolution 
(Section~\ref{sec:rr}), we use several combinations of distributional information and lexical resources
to build a \emph{lexical and phrasal entailment rule classifier} (\emph{entailment rule classifier} for short)
for weighting the rules
appropriately. These extracted rules create an especially valuable
resource for testing lexical entailment systems, as they contain a
variety of entailment relations (hypernymy, synonymy, antonymy, etc.),
and are actually useful in an end-to-end RTE system.

We describe the entailment rule classifier in multiple parts. In
Section~\ref{sec:entlex}, we overview a lexical entailment rule classifier,
which only handles single words. Section~\ref{sec:entresources} describes
the lexical resources used.
In Section~\ref{sec:entasym}, we describe how our
previous work in supervised hypernymy detection is used in the system.
In Section~\ref{sec:entphrasal}, we describe the approaches for extending
the classifier to handle phrases.

%% Finally,
%% in Section~\label{sec:entexperiment}, we describe the experimental procedure
%% and details of the machine learning components.

\subsubsection{Lexical Entailment Rule Classifier}
\label{sec:entlex}

We begin by describing the lexical entailment rule classifier, which
only predicts entailment between single words, treating the task as a
supervised classification problem given the lexical rules  
constructed from the modified Robinson resolution as input.
We use numerous features which we expect to be predictive of lexical
entailment.  Many were previously shown to be successful for
the SemEval 2014 Shared Task on lexical entailment
\cite{marelli:semeval14,bjerva:2014semeval,lai:semeval14}.  Altogether,
we use four major groups of features 
as  summarized in
Table~\ref{tab:lexentfeat} and described in 
detail below.

\begin{table}
\centering
 \footnotesize 
\begin{tabular}{|lllr|}
    \hline
    \bf{Name} & \bf{Description} & \bf{Type} & \bf{\#}\\
    \hline\hline
    \multicolumn{3}{|l}{\bf{Wordform}} & 18\\
    \hline
    Same word & Same lemma, surface form & Binary & 2\\
    POS & POS of LHS, POS of RHS, same POS & Binary & 10\\
    Sg/Pl & Whether LHS/RHS/both are singular/plural & Binary & 6\\
    \hline
    \hline
    \multicolumn{3}{|l}{\bf WordNet} & 18\\
    \hline
    OOV & True if a lemma is not in WordNet, or no path exists & Binary & 1\\
    Hyper & True if LHS is hypernym of RHS & Binary & 1\\
    Hypo & True if RHS is hypernym of LHS & Binary & 1\\
    Syn & True if LHS and RHS is in same synset & Binary & 1\\
    Ant & True if LHS and RHS are antonyms & Binary & 1\\
    Path Sim & Path similarity (NLTK) & Real & 1\\
    Path Sim Hist & Bins of path similarity (NLTK) & Binary & 12\\
    \hline
    \hline
    \multicolumn{3}{|l}{\bf Distributional features (Lexical)} & 28\\
    \hline
    OOV & True if either lemma not in dist space & Binary & 2\\
    BoW Cosine & Cosine between LHS and RHS in BoW space & Real & 1\\
    Dep Cosine & Cosine between LHS and RHS in Dep space & Real & 1\\
    BoW Hist & Bins of BoW Cosine & Binary & 12\\
    Dep Hist & Bins of Dep Cosine & Binary & 12\\
    \hline
    \hline
    \multicolumn{3}{|l}{\bf Asymmetric Features \cite{roller:coling14}} & 600\\
    \hline
    Diff   & LHS dep vector $-$ RHS dep vector & Real & 300\\
    DiffSq & RHS dep vector $-$ RHS dep vector, squared & Real & 300\\
    \hline
\end{tabular}
\caption{List of features in the lexical entailment classifier, along
with types and counts}
\label{tab:lexentfeat}
\end{table}

{\em Wordform Features} We extract a number of simple features based on
the usage of the LHS and RHS in their original sentences. We extract
features for whether the LHS and RHS have the same lemma, same surface form,
same POS, which POS tags they have, and whether they are singular or plural.
Plurality is determined from the POS tags.

{\em WordNet Features} We use WordNet 3.0 to determine whether the LHS
and RHS have known synonymy, antonymy, hypernymy, or hyponymy relations. We
disambiguate between multiple synsets for a lemma by selecting the synsets for
the LHS and RHS which minimize their path distance. If no path exists, we
choose the most common synset for the lemma. Path similarity, as implemented
in the Natural Language Toolkit~\cite{bird2009natural}, is also used as a feature.

{\em Distributional Features} We measure distributional %semantic
similarity in two distributional spaces, one which models topical
similarity (BoW), and one which models syntactic similarity (Dep).
We use cosine similarity of the LHS and RHS in both spaces as
features.

\begin{figure}
    \begin{center}
    \includegraphics[width=0.8\textwidth]{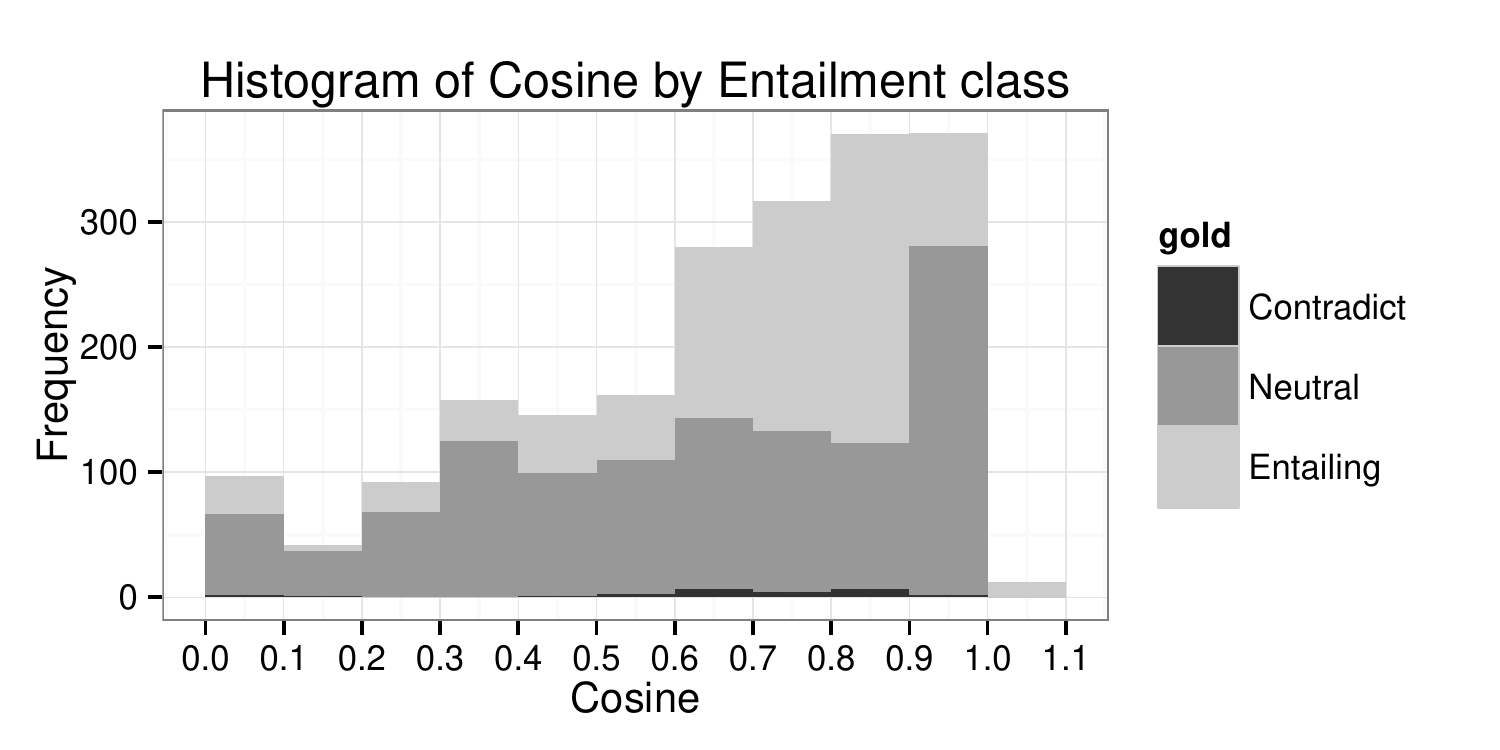}
    \end{center}
\caption{Distribution of entailment relations on lexical items by cosine.
Highly similar pairs (0.90-0.99) are less likely entailing than moderately
similar pairs (0.70-0.89).}
\label{fig:valley}
\end{figure}

One very important feature set used from distributional similarity is
the {\em histogram binning} of the cosines. We create 12 additional binary,
mutually-exclusive features, which mark whether the distributional similarity
is within a given range. We use the ranges of exactly 0, exactly 1, 0.01-0.09,
0.10-0.19, \dots, 0.90-0.99. Figure \ref{fig:valley} shows the importance of
these histogram features: words that are very similar (0.90-0.99) are much
less likely to be entailing than words which are moderately similar
(0.70-0.89). This is because the most highly similar words are likely
to be cohyponyms. 
% to be in co-hyponymy than hypernymy or other entailing relationships.

\subsubsection{Preparing Distributional Spaces}
\label{sec:entresources}

As described in the previous section, we use distributional semantic similarity
as features for the entailment rules classifier. Here we describe the preprocessing steps to
create these distributional resources.

{\em Corpus and Preprocessing:} We use the BNC, ukWaC and a 2014-01-07 copy
of Wikipedia. All corpora are preprocessed %tokenized, POS tagged, lemmatized, and dependency parsed
using Stanford CoreNLP. We collapse particle verbs into a single
token, and all tokens are annotated with a (short) POS tag
so that the same lemma with a different POS is modeled separately. We keep only
content words (NN, VB, RB, JJ) appearing at least 1000 times in the corpus. The
final corpus contains 50,984 types and roughly 1.5B tokens.
%create distributional representations for content lemmas (NN, VB, JJ, RB)% verbs,
%adjectives, and adverbs)
%appearing at least 1000 times in the concatenated
%corpus.
%The final distributional spaces model 50,984 unique lemma/POS types
%and are based on roughly 1.5B tokens.

{\em Bag-of-Words vectors:} We filter all but the 51k chosen lemmas from
the corpus, and create one sentence per line. We use Skip-Gram Negative Sampling to create vectors \cite{mikolov:2013iclr}. We use 300
latent dimensions, a window size of 20, and 15
negative samples. These parameters were not tuned, but chosen as
reasonable defaults for the task. We use the large window
size to ensure the BoW vectors captured more topical %and general word
similarity, rather than syntactic similarity, which is modeled by %explicitly by
the dependency vectors.

{\em Dependency vectors:} We extract {\em (lemma/POS, relation, context/POS)}
tuples from each of the Stanford Collapsed CC Dependency graphs. We filter tuples with lemmas not in our 51k chosen types.
Following \namecite{typeDM}, we model inverse relations and mark them separately. For example,
%Keeping in line with prior work in syntactic distributional spaces~\cite{typeDM},
%we model inverse relations as well, but mark them separately. For example,
``red/JJ car/NN'' will generate tuples for both {\em (car/NN, amod, red/JJ)} and
{\em (red/JJ, amod$^{-1}$, car/NN)}. After extracting tuples, we discard all
but the top 100k {\em (relation, context/POS)} pairs and build a vector space using
{\em lemma/POS} as rows, and {\em (relation, context/POS)} as columns. The matrix
is transformed with Positive Pointwise Mutual Information (PPMI), and reduced to
300 dimensions using Singular Value Decomposition (SVD). We do not vary these
parameters, but chose them as they performed best in prior work \cite{roller:coling14}. %as they performed best in previous work on identifying
%lexical relations \cite{roller:coling14}.

\subsubsection{Asymmetric Entailment Features}
\label{sec:entasym}

As an additional set of features, we also use the representation
previously employed by the asymmetric, supervised
hypernymy classifier described by \namecite{roller:coling14}. Previously,
this classifier was only used on artificial datasets, which encoded specific
lexical relations, like hypernymy, co-hyponymy, and meronymy. Here, we use
its representation to encode just the three general relations: entailment, neutral, and
contradiction.

The asymmetric features take inspiration from \namecite{mikolov:2013naacl}, who
found that differences between distributional vectors often encode certain linguistic regularities,
like $\vec{king} - \vec{man} + \vec{woman} \approx \vec{queen}$. In particular
the asymmetric classifier uses two sets of features, $<f, g>$, where:
\begin{align*}
  f_i(LHS, RHS) & = \vec{LHS}_i - \vec{RHS}_i\\
  g_i(LHS, RHS) & = f_i^2,
\end{align*}
that is, the vector difference between the LHS and the RHS, and this difference
vector squared. Both feature sets are extremely important to strong performance.

For these asymmetric features, we use the Dependency space described
earlier.  We choose the Dep space because we previously found that spaces
reduced using SVD outperform word embeddings generated by the Skip-gram
procedure. We do not use both spaces, because of the large number of features
this creates.

Recently, there have been considerable work in detecting lexical entailments
using only distributional vectors. The classifiers proposed by
\namecite{fu:acl:2014}; \namecite{levy:2015tt}; and \namecite{kruszewski2015deriving}
could have also been used in place of these asymmetric features, but we reserve
evaluations of these models for future work.

\subsubsection{Extending Lexical Entailment to Phrases}
\label{sec:entphrasal}

The lexical entailment rule classifier described in previous sections is limited
to only simple rules, where the LHS and RHS are both single words. Many of
the rules generated by the modified Robinson resolution are actually {\em phrasal} rules,
such as {\em little boy $\rightarrow$ child}, or
{\em running $\rightarrow$ moving quickly}. In order to model these phrases, we use two general approaches: first, we
use a state-of-the-art compositional model, in order
to create vector representations of phrases, and then include the same
similarity features described in the previous section. The
full details of the compositional distributional model are described
in Section~\ref{sec:pengxiang}. 

In addition to a compositional distributional model, we also used a simple, greedy
word aligner, similar to the one described by \namecite{lai:semeval14}. This
aligner works by finding the pair of words on the LHS and RHS which are
most similar in a distributional space, and marking them as ``aligned''.
The process is repeated until at least one side is completely exhausted. For
example, ``red truck $\rightarrow$ big blue car'', we would align ``truck'' with
``car'' first, then ``red'' with ``blue'', leaving ``big'' unaligned.
%The alignment algorithm is listed in Algorithm~\ref{algo:greedyalign}.

% KE: saving space; we can put this back in if we want to.
% \begin{algorithm}[!t]
% \caption{Greedy Alignment of Lexical Items}
% \label{algo:greedyalign}
% \begin{algorithmic}[1]
%     \Require $D$: word $\rightarrow R^{k}$, a distributional vector space
%     \Require $L$: a set of words on the LHS
%     \Require $R$: a set of words on the RHS
%     \Ensure $A$: a set of aligned word pairs
%     \Ensure $L', R'$: unaligned words on LHS, RHS
%     \State $A := \{\}$
%     \While{$||L|| > 0$ and $||R|| > 0$}
%         \State Select $l \in L$, $r \in R$ which maximizes $cosine(\vec{l}, \vec{r})$
%         \State Add $(l, r)$ to $A$
%         \State Remove $l$, $r$ from $L$, $R$ respectively
%     \EndWhile
%     \State \Return A, L, R
% \end{algorithmic}
% \end{algorithm}

After performing the phrasal alignment, we compute a number of {\em base}
features, based on the results of the alignment procedure. These
include values like the length of the rule, the percent of words unaligned,
etc. We also compute all of the same features used in the lexical
entailment rule classifier (Wordform, WordNet, Distributional) and compute
their min/mean/max across all the alignments. We do not include the
asymmetric entailment features as the feature space then becomes
extremely large. Table~\ref{tab:phrasalentfeat} contains
a listing of all phrasal features used.

\begin{table}
\centering
 \footnotesize 
\begin{tabular}{|lllr|}
    \hline
    \bf{Name} & \bf{Description} & \bf{Type} & \bf{\#}\\
    \hline\hline
    \multicolumn{3}{|l}{\bf Base} & 9\\
    \hline
    Length & Length of rules & Real & 2\\
    Length Diff & Length of LHS - length of RHS & Real & 1\\
    Aligned & Number of alignments & Real & 1\\
    Unaligned & Number of unaligned words on LHS, RHS & Real & 2\\
    Pct aligned & Percentage of words aligned & Real & 1\\
    Pct unaligned & Percentage of words unaligned on LHS, RHS & Real & 2\\
    \hline
    \hline
    \multicolumn{3}{|l}{\bf Distributional features \cite{paperno:acl14}} & 16\\
    \hline
    Cosine & Cosine between mean constituent vectors & Real & 1\\
    Hist   & Bins of cosine between mean constituent vectors & Binary & 12\\
    Stats  & Min/mean/max between constituent vectors & Real & 3\\
    \hline
    \hline
    \multicolumn{3}{|l}{\bf Lexical features of aligned words} & 192\\
    \hline
    Wordform & Min/mean/max of each Wordform feature &  & 54\\
    WordNet & Min/mean/max of each WordNet feature &  & 54\\
    Distributional & Min/mean/max of each Distributional feature &  & 84\\
    \hline
\end{tabular}
\caption{Features used in Phrasal Entailment Classifier, along with types and counts.}
\label{tab:phrasalentfeat}
\end{table}

\subsubsection{Phrasal Distributional Semantics}
\label{sec:pengxiang}

We build phrasal distributional space based on the practical lexical function model of \namecite{paperno:acl14}. We again use as the corpus a concatenation of BNC, ukWaC and English Wikipedia, parsed with the Stanford CoreNLP parser. We focus on 5 types of dependency labels, ``amod'', ``nsubj'', ``dobj'', ``pobj'', ``acomp'', and combine the governor and dependent words of these dependencies to form adjective-noun, subject-verb, verb-object, preposition-noun and verb-complement phrases respectively.
We only retain phrases where both the governor and the dependent are among the 50K most frequent words in the corpus, resulting in 1.9 million unique phrases. The co-occurrence counts of the 1.9 million phrases with the 20K most frequent neighbor words in a 2-word window are converted to a PPMI matrix, and reduced to 300 dimensions by performing SVD on a lexical space and applying the resulting representation to the phrase vectors, normalized to length 1.

Paperno et al.\ represent a word as a vector, which
represents the contexts in which the word can appear, along with a
number of matrices, one for each type of dependent that the word can
take. For a transitive verb like \textit{chase}, this would be one
matrix for subjects, and one for direct objects. The representation of
the phrase \textit{chases dog} is then 
\[\vec{chase} + \mathop{chase}^{\Box_o} \times \vec{dog}\]
where $\times$ is matrix multiplication, and when the phrase is
extended with \textit{cat} to form \textit{cat
  chases dog}, the representation is
\[\vec{chase} + \mathop{chase}^{\Box_s} \times \vec{cat} + (\vec{chase} + \mathop{chase}^{\Box_o} \times \vec{dog})\]
For verbs, the practical lexical function model trains a matrix for each of the relations \textit{nsubj}, \textit{dobj} and \textit{acomp}, for adjectives a matrix for \textit{amod}, and for prepositions a matrix for \textit{pobj}. For example, 
the \textit{amod} matrix of the adjective ``red/JJ'' is trained as follows. We collect all phrases in which ``red/JJ'' serves as adjective modifier (assuming the number of such phrases is $N$), like ``red/JJ car/NN'', ``red/JJ house/NN'' etc., and construct two $300\times N$ matrices $M_{arg}$ and $M_{ph}$, where the $i$th column of $M_{arg}$ is the vector of the noun modified by ``red/JJ'' in the $i$th phrase ($\overrightarrow{car}, \overrightarrow{house}$, etc.), and the $i$th column of $M_{ph}$ is vector of phrase $i$ minus the vector of ``red/JJ'' ($\overrightarrow{red\ car} - \overrightarrow{red}, \overrightarrow{red\ house} - \overrightarrow{red}$, etc.), normalized to length 1.
Then the \textit{amod} matrix $\mathop{red}^{\Box(amod)}\in R^{300\times 300}$ of ``red/JJ'' can be computed via ridge regression.
Given trained matrices, we compute the composition vectors by applying the functions recursively starting from the lowest dependency. 
%  as:
% \begin{equation*}
% \mathop{red}^{\Box(amod)} = {\arg\min}_{X}\Vert M_{Ph} - X\times M_{arg}\Vert_2^2 + \lambda\Vert X\Vert_2^2
% \end{equation*}
% As the dimension of $X$ is fairly small, the ridge regression can be efficiently solved with standard solution as:
% \begin{equation*}
% \mathop{red}^{\Box(amod)} = M_{ph}  M_{arg}^T  (M_{arg} M_{arg}^T+\lambda I)^{-1}
% \end{equation*}
%The ridge parameter $\lambda$ is selected from \{0.1,1,10\} by generalized cross validation.

As discussed above, some of the logical rules from Section \ref{sec:rr} need to be split into multiple rules. We use the dependency parse to split long rules by iteratively searching for the highest nodes in the dependency tree that occur in the logical rule, and identifying the logical rule words that are its descendants in phrases that the practical lexical functional model can handle. % For example the sentence \textit``{A girl with a black bag is on a crowded train''} yields gives the logical rule as \textit{``girl with bag on crowded train''}, which we cannot get one single vector as the verb ``is'' is not in the rule. So the rule is split into two short phrases \textit{``girl with bag''} and \textit{``on crowded train''}.
After splitting, we perform greedy alignment on phrasal vectors to pair up rule parts. Similar to Section \ref{sec:entphrasal}, we iteratively identify the pair of phrasal vectors on the LHS and RHS which have the highest cosine similarity until one side has no more phrases.

\subsection{Precompiled Rules}
\label{sec:precompiled}

The second group of rules is collected from existing databases. 
We collect rules from WordNet~\cite{princeton:url10} and the  paraphrase collection PPDB~\cite{ganitkevitch:naacl13}. 
We use simple string matching to find the set of rules that are relevant to 
a given text/query pair $T$ and $H$. If the left-hand side of a rule is a substring of $T$ and the right-hand is 
a substring of $H$, the rule is added, and likewise for rules with LHS
in $H$ and RHS in $T$. Rules that go from $H$ to $T$ 
are important in case $T$ and $H$ are negated, e.g. 
$T$: \textit{No ogre likes a princess},
$H$: \textit{No ogre loves a princess}. 
The rule needed is \textit{love $\Rightarrow$ like} which goes from $H$ to $T$. 

\paragraph{WordNet}
\label{sec:wn}
WordNet~\cite{princeton:url10} is a lexical database of words grouped into 
sets of synonyms. In addition to grouping synonyms, it lists semantic 
relations connecting groups. We represent the information on WordNet as 
``hard'' logical rules. The semantic relations 
we use are: 
\begin{itemize}
\item Synonymy: $\forall x. \: man(x) \Leftrightarrow guy(x) $
\item Hypernymy:  $\forall x. \: car (x) \Rightarrow vehicle(x) $
\item Antonymy: $\forall x. \: tall(x) \Leftrightarrow \lnot short(x)$
\end{itemize}
One advantage of using logic is that it is a powerful 
representation that can effectively represent these different semantic relations.

\paragraph{Paraphrase collections}
\label{sec:ppdb}
Paraphrase collections are precompiled sets of rules, e.g: 
\textit{a person riding a bike $\Rightarrow$ a biker}. We translate paraphrase 
collections, in this case PPDB~\cite{ganitkevitch:naacl13}, to logical rules.
We use the Lexical, One-To-Many and Phrasal sections of the XL version
of PPDB.  

We use a simple rule-based approach to translate natural-language
rules to logic. First, we can make the assumption that the translation
of a PPDB rule is going to be a conjunction of positive atoms. PPDB
does contain some rules that are centrally about negation, such as
\textit{deselected $\Rightarrow$ not selected}, but we skip those as
the logical form analysis already handles negation. As always, we want
to include in $KB$ only rules pertaining to a particular text/query
pair $T$ and $H$. Say $LHS \Rightarrow
RHS$ is a rule such that $LHS$ is a substring of $T$ and $RHS$ is a
substring of $H$. Then each word in $LHS$ gets represented by a unary
predicate applied to a variable, and likewise for $RHS$ -- note
that we can expect the same predicates to appear in the logical forms
$L(T)$ and $L(H)$ of the text and query. For example, if the rule is  \textit{a person
  riding a bike $\Rightarrow$ a biker}, then we get the atoms $person(p)$,
$riding(r)$ and $bike(b)$ for the $LHS$, with variables $p, r, b$. We then
add Boxer meta-predicates to the logical form for $LHS$, and likewise
for $RHS$. Say that $L(T)$ includes $person(A) \wedge ride(B) \wedge
bike(C) \wedge agent(B, A) \wedge patient (B, C)$ for constants $A$,
$B$, and $C$, then we extend the logical form for $LHS$ with $agent(r,
p) \wedge patient(r, b)$. We
proceed analogously for $RHS$. This gives us the logical forms:
$L(LHS) = person(p) \wedge agent(r, p)  \wedge  riding(r) \wedge   patient(r, b) \wedge bike(b)$
and $L(RHS) = biker(k)$. 

The next step is to bind the variables in $L(LHS)$ to those in
$L(RHS)$.  In the example above, the variable $k$ in the $RHS$ should be
matched with the variable $p$ in the $LHS$. We determine these bindings using a
simple rule-based approach: We manually define paraphrase rule
\emph{templates} for PPDB, which specify variable bindings. A rule template is
conditioned on the part-of-speech
tags of the words involved. In our example it is $N_1V_2N_3
\Rightarrow N_1$, which binds the variables of the first $N$ on the left to the first $N$ on
the right, unifying the variables $p$ and $k$.  The final
paraphrase rule is: $\forall p, r, b. \: person(p) \wedge agent(r, p) \wedge riding(r)
\wedge patient(r, b) \wedge bike(b) \Rightarrow biker(p)$.  In case some
variables in the $RHS$ remain unbound, they are
existentially quantified, e.g.: $\forall p. \: pizza(p) \Rightarrow \exists
q. \: slice(p) \wedge of (p, q) \wedge pizza (q) $.

%\paragraph{Weight Mapping}
Each PPDB rule comes with a set of similarity scores which
we need to map to a single MLN weight. We use the simple log-linear equation suggested by
\namecite{ganitkevitch:naacl13} to map the scores into a single value:
\begin{equation}
weight(r) = - \sum^{N}_{i=1} \lambda_i \log  \varphi_i
\end{equation}
where, $r$ is the rule, $N$ is number of the similarity scores provided for the rule $r$, 
$\varphi_i$ is the value of the $i$th score, and $\lambda_i$ is its scaling factor. 
For simplicity, following ~\namecite{ganitkevitch:naacl13}, we set all $\lambda_i$ to 1.
To map this weight to a final MLN rule weight, 
we use the weight-learning method discussed in Section~\ref{sec:wlearn}.

\paragraph{Handcoded rules}
\label{sec:hand}
We also add a few handcoded rules to the $KB$ that we do not get from other resources. 
For the SICK dataset, we only add several lexical rules where one side of the rule 
is the word \textit{nobody}, e.g:  \textit{nobody $\Leftrightarrow \lnot$ somebody}
and \textit{nobody $\Leftrightarrow \lnot$ person}.

\section{Probabilistic Logical Inference}
\label{sec:infer}
We now turn to the last of the three main components of our system,
probabilistic logical inference. MLN inference is usually intractable,
and using MLN 
implementations ``out of the box'' 
does not work for our application. This section discusses an MLN implementation 
that supports complex queries and uses the closed
world assumption (CWA) to decrease problem size, hence making inference more efficient. Finally, 
this section discusses a simple weight learning scheme to learn global
scaling factors for weighted rules in $KB$ from different sources. 

%Other than weight learning, this section mostly draws on 
%the previous paper by \namecite{beltagy:starai14}.

\subsection{Complex formulas as queries}
\label{sec:qf}
Current implementations of MLNs like Alchemy~\cite{kok:tr05} do not allow 
queries to be complex formulas, they can only calculate probabilities of 
ground atoms. This section discusses an inference algorithm 
for arbitrary query formulas.

\paragraph*{The standard work-around}
\label{sec:workaround}
Although current MLN implementations can only calculate probabilities
of ground atoms, they can be used to calculate the probability of a
complex formula through a simple work-around. The complex query formula $H$
is added to the MLN using the hard formula:
\begin{equation}
\label{eq:workaround}
H \Leftrightarrow result(D) \: | \: \infty
\end{equation}
where $result(D)$ is a new ground atom that is not used anywhere else in the MLN. 
Then, inference is run to calculate the probability of $result(D)$, 
which is equal to the probability of the formula $H$. 
%%DELETED% summerized below
%However, this 
%approach can be very inefficient for some queries. 
%For example, consider the following query:
%\begin{align}
%\label{eq:example}
%H: \:  \exists x,y,z.  \: ogre(x) \wedge agent(y,x) \wedge love(y) 
% \wedge patient(y,z) \wedge princess(z)
%\end{align}
%This form of an existentially quantified formula with a list of
%conjunctively joined atoms is very common in the inference problems
%we are addressing, so it is important to have efficient inference for
%such queries.  In general, logical forms of natural language sentences
%usually comprise large numbers of existentially quantified variables
%and only rarely any other quantifiers. However, using this particular  $H$ in Equation~\ref{eq:workaround}
%results in a very inefficient MLN. For the direction $H\Leftarrow
%result(D) $ of the double-implication in Equation~\ref{eq:workaround},
%the existentially
%quantified formula is replaced with a large disjunction over all
%possible combinations of constants for variables $x,y$ and $z$~\cite{gogate:uai11}. Generating this disjunction, converting it to
%clausal form, and running inference on the resulting ground network
%becomes increasingly intractable as the number of variables and
%constants grow.
However, this 
approach can be very inefficient for the most common form of queries, which are existentially quantified 
queries, e.g: 
\begin{align}
\label{eq:example}
H: \:  \exists x,y,z.  \: ogre(x) \wedge agent(y,x) \wedge love(y) 
 \wedge patient(y,z) \wedge princess(z)
\end{align}
Grounding  of the backward direction of the double-implication is very problematic 
because the existentially quantified formula is replaced with a large disjunction over all
possible combinations of constants for variables $x,y$ and $z$~\cite{gogate:uai11}. Converting this disjunction to
clausal form becomes increasingly intractable as the number of variables and constants grow.

\paragraph*{New inference method}
Instead, we propose an inference algorithm to directly calculate the
probability of complex query formulas.  In MLNs, the probability of a
formula is the sum of the probabilities of the possible worlds that
satisfy it. \namecite{gogate:uai11} show that to
calculate the probability of a formula $H$ given a probabilistic
knowledge base $KB$, it is enough to compute the partition function $Z$
of $KB$ with and without $H$ added as a hard formula:
\begin{equation}
P ( H \mid KB ) = \frac{Z(KB \cup \{(H,\infty)\})}{Z(KB)}
\end{equation}
Therefore, all we need is an appropriate algorithm to estimate 
the partition function $Z$ of a Markov network. Then, we construct 
two ground networks, one with the query and one without, and
estimate their $Z$s  using that estimator. The ratio 
between the two $Z$s is the probability of $H$.

%DELETED% for space constraint 
%We tried to estimate $Z$ using a harmonic-mean estimator on the samples
%generated by MC-SAT~\cite{poon:aaai06}, a popular and generally effective MLN
%inference algorithm, but we found that the estimates are highly inaccurate as
%shown by \namecite{venugopal:aaai13}.  Instead we use 
We use
SampleSearch~\cite{gogate:ai11} to estimate the partition
function. SampleSearch is an importance sampling algorithm that has been shown
to be effective when there is a mix of probabilistic and
deterministic (hard) constraints, a fundamental property of the inference
problems we address.  Importance sampling in general is problematic in the
presence of determinism, because many of the generated samples violate the
deterministic constraints, and they get rejected. Instead, SampleSearch uses a
base sampler to generate samples then uses backtracking search with a SAT
solver to modify the generated sample if it violates the deterministic
constraints. We use an implementation of SampleSearch that uses a generalized
belief propagation algorithm called Iterative Join-Graph Propagation
(IJGP)~\cite{dechter:uai02} as a base sampler.  This version is available
online~\cite{gogate:online}.

For cases like the example $H$ in Equation~\ref{eq:example}, we need
to avoid
generating a large disjunction because of the existentially quantified
variables. So we replace $H$ with its negation $\ \lnot H$, replacing
the existential quantifiers with universals, which are easier
to ground and perform inference upon.  Finally, we compute the
probability of the query $P(H) = 1- P(\lnot H)$. Note that replacing 
$H$ with $\ \lnot H$ cannot make inference with the standard work-around faster, 
because with $\ \lnot H$, the direction
$ \lnot H \Rightarrow result(D) $ suffers from the same problem 
of existential quantifiers that we previously had with 
$  H \Leftarrow result(D) $.

\subsection{Inference Optimization using the Closed-World Assumption}
\label{sec:mcw}
This section explains why our MLN inference problems are computationally
difficult, then explains how the closed-world assumption (CWA) can be used to reduce the problem size and
speed up inference. For more details, see \namecite{beltagy:starai14}.

%In Section~\ref{sec:parse} we have characterized the probabilistic
%entailment of a query from a text through the conditional probability $P(H |T, KB, W_{T,H})$, where $H$ is the query,
%$T$ is the text, $KB$ is the set of rules, and $W_{T,H}$ is the world configuration. 
%An important distinction for practical implementations
%of Markov logic networks is between the
% \emph{evidence set}, a set of weighted ground literals, and the
% \emph{rules}, weighted first-order formulas that are not just weighted
% ground literals. If we write $E$ for the evidence set and $R$ for the
% set of rules, we can reformulate our probabilistic inference as
%$P(H|E, R, W)$. We use this formulation for what follows.
%
In the inference problems we address, formulas are typically long, especially
the query $H$. The number of ground clauses of a first-order
formula is exponential in the number of variables in the formula, it
is $O(c^v)$,
where $c$ is number of constants in the domain and $v$ is number of variables
in the formula.  For a moderately long formula, the number of resulting
ground clauses is infeasible to process.

We have argued above (Section~\ref{sec:cwa}) that for probabilistic
inference problems based on natural language text/query pairs, it
makes sense to make the closed world assumption: If we want to know if
the query is true in the situation or setting laid out in the text, we
should take as false anything not said in the text. In our
probabilistic setting, the CWA amounts to giving low prior
probabilities to all ground atoms unless they can be inferred
from the text and knowledge base. 
However, we found that a large fraction
of the ground atoms cannot be inferred from the text and knowledge
base, and their probabilities remain very
low.  
As an approximation, we can assume that this small probability is exactly zero and these ground atoms are false, 
without significantly affecting the probability of the query. 
%This suggests that these ground atoms can be identified and removed in
%advance with very little impact on the approximate nature of the inference.  
%As the number of such ground atoms is large, this has the potential to
This will remove a large number of the ground atoms, which will 
dramatically decrease the size of the ground network and speed up inference.

We assume that all ground atoms are false by default unless they are can be inferred from
the text and the knowledge base $T \wedge KB$. 
For example:
\begin{equation*}
\begin{aligned}
T: \: &ogre(O) \wedge agent(S,O) \wedge snore(S) \\
KB: \: &\forall x. \: ogre(x) \Rightarrow monster(x)  \\
H: \: &\exists x,y. \: monster(x) \wedge agent(y, x) \wedge snore(y)
 \end{aligned}
\end{equation*}
Ground atoms $\{ogre(O), snore(S), agent(S,O)\}$  are not false because they can be  inferred from  $T$. 
Ground atom $monster(O)$ is also not false 
because it can be inferred from $T \wedge KB$. All other ground atoms are false. 

Here is an example of how this simplifies the query $H$. 
$H$ is equivalent to a disjunction of all its possible groundings:
$H: \:  (monster(O) \wedge agent(S, O) \wedge snore(S))
\vee (monster(O) \wedge agent(O, O) \wedge snore(O))
\vee (monster(S) \wedge agent(O, S) \wedge snore(O))
\vee (monster(S) \wedge agent(S, S) \wedge snore(S))$. 
Setting all ground atoms to false except the inferred ones, then simplifying the expression, we get:
$H: \:  monster(O) \wedge agent(S, O) \wedge snore(S)$.
Notice that most ground clauses of $H$ are removed because they are False. 
We are left just with the ground clauses that potentially have a non-zero probability.
Dropping all False ground clauses leaves an exponentially smaller number of ground clauses 
in the ground network. Even though the inference
problem remains exponential in principle, the problem is much smaller
in practice, such that inference becomes feasible. 
In our experiments with the SICK dataset, the number of ground clauses 
for the query ranges from 0 to 19,209 with mean 6.
This shows that the CWA effectively reduces the number of ground clauses for the query 
from millions (or even billions) to a manageable number. 
With the CWA, the average number of inferrable ground atoms 
(ignoring ground atoms from the text) ranges from 0 to 245 with an average of 18.

\subsection{Weight Learning}
\label{sec:wlearn}

We use weighted rules from different sources, both PPDB weights
(Section \ref{sec:ppdb}) and the confidence of the entailments rule
classifier (Section \ref{sec:ent}). These weights are not necessarily
on the same scale, for example one source could produce systematically
larger weights than the other. To map them into uniform weights that
can be used within an MLN, we use 
weight learning.
Similar to the work of \namecite{zirn:ijcnlp11}, 
we learn a single mapping parameter for each source of rules that functions as
a scaling factor: 
\begin{equation}
\label{eq:wlearn}
MLNweight = scalingFactor \times ruleWeight
\end{equation}
We use a simple grid search to learn the scaling factors that optimize
performance on the RTE training data.

Assuming that all rule weights are in  $[0, 1]$
(this is the case for classification confidence scores, and PPDB weights can be scaled), 
we also try the following mapping:
\begin{equation}
\label{eq:wlearnlog}
MLNweight = scalingFactor \times \log( \frac{ruleWeight}{1-ruleWeight})
\end{equation}
This function assures that for an MLN with a single rule 
$LHS \Rightarrow RHS \: | \: MLNweight$, 
it is the case that 
$P(RHS|LHS) = ruleWeight$, 
given that $scalingFactor = 1$.

\section{Evaluation}
\label{sec:eval}
This section evaluates our system. First, we 
evaluate several lexical and phrasal distributional systems 
on the  rules that we collected using modified Robinson resolution.
This includes an in-depth analysis of different types of
distributional information within the entailment rule classifier.
Second, we use the best configuration we find in the first step 
as a knowledge base and evaluate our system on the RTE task using the SICK dataset. 
%We evaluate different components of the system and study the impact of each one on the final system's accuracy. 

{\em Dataset:}
The SICK dataset, which is described in Section~\ref{sec:relwork},
consists of 5,000 pairs for training and 4,927
for testing. Pairs are annotated for RTE and STS (Semantic Textual
Similarity) tasks.  We use the RTE annotations of the dataset.

\subsection{Evaluating the Entailment Rule Classifier}
\label{sec:lexeval}
The entailment rule classifier described in Section~\ref{sec:ent} constitutes a
large portion of the full system's end-to-end performance, but consists of many
different feature sets providing different kinds of information. In this
section, we thoroughly evaluate the entailment rule classifier, both
quantitatively and qualitatively, to identify the individual and holistic value
of each feature set and systematic patterns. However, this evaluation may
also be used as a framework by future lexical semantics research to see its
value in end-to-end textual entailment systems. For example, we could have
also included features corresponding to the many measures of distributional
inclusion which were developed to predict hypernymy
\cite{weedsweirmccarthy:2004:COLING,kotlerman:nlej10,lenci:starsem12,santus:2013},
or other supervised lexical entailment classifiers \cite{Baroni:ut,fu:acl:2014,weeds:coling:2014,levy:2015tt,kruszewski2015deriving}.

%We evaluate the entailment rule classifier described in Section~\ref{sec:ent}.
Evaluation is broken into four parts: first, we overview performance of the entire
entailment rule classifier on all rules, both lexical and phrasal. We then break
down these results into performance on only lexical rules and only
phrasal rules. Finally, we look at only the asymmetric features to
address concerns raised by \namecite{levy:2015tt}. In all sections, we evaluate
the lexical rule classifier on its ability to generalize to new word pairs, as
well as the full system's performance when the entailment rule
classifier is used as the {\em only} source of knowledge.
% We also perform some qualitative
% analysis in order to highlight the strengths and weaknesses of particular
% feature sets.

Overall, we find that distributional semantics is of vital importance to the
lexical rule classifier and the end-to-end system, especially when word
relations are not explicitly found in WordNet.  The introduction of syntactic
distributional spaces and {\em cosine binning} are especially valuable, and
greatly improve performance over our own prior work. Contrary to
\namecite{levy:2015tt}, we find the asymmetric features provide better
detection of hypernymy over memorizing of prototypical hypernyms, but the prototype vectors
better capture examples which occur very often in the data;  explicitly
using both does best. Finally, we find, to our surprise, that a 
state-of-the-art compositional distributional method \cite{paperno:acl14}
yields disappointing performance on phrasal entailment detection, though it
does successfully identify non-entailments deriving from
changing prepositions or semantic roles.

\subsubsection{Experimental Setup}
We use the gold standard annotations described in Section~\ref{sec:annotaterules}.
We perform 10 fold cross-validation on the annotated training set, using the same
folds in all settings. Since some RTE sentence pairs require multiple lexical
rules, we ensure that cross-validation folds are stratified across the sentences, so that
the same sentence cannot appear in both training and testing. We use a Logistic Regression classifier with an L2 regularizer.\footnote{We experimented with
  multiple classifiers, including Logistic Regression, 
Decision Trees, and SVMs (with polynomial, RBF, and linear kernels). We found
that linear classifiers, and chose Logistic
Regression, since it was used in \namecite{roller:coling14} and \namecite{lai:semeval14}.}
%We also found an L2 regularizer slightly outperformed an L1 regularizer.
%It is not immediately obvious why a linear classifier should perform best on
%the task, but we believe it is because a great deal of nonlinearity is already
%encoded in the features themselves. For example, {\em Same POS} and the {\em
%Histogram} and {\em Min/Mean/Max} features all capture simple nonlinearities,
%without giving the classifier freedom to search for interaction between
%unrelated variables. It is worth noting, however, that \namecite{lai:semeval14}
%use similar features and also found success with a Logistic Regression
%classifier.}
Since we perform
three-way classification, we train models using one-vs-all.

Performance is measured in two main metrics. \textit{Intrinsic
  accuracy} measures how
the classifier performs in the cross-validation setting on the
training data. This corresponds to
treating lexical and phrasal entailment as a basic supervised learning
problem. \emph{RTE accuracy} is accuracy on the end task of
textual entailment using the predictions of the entailment rule
classifier. For RTE accuracy, the predictions of the
entailment rule classifier were used as the only knowledge base in the RTE
system. RTE training accuracy uses the predictions from the
cross-validation experiment, and for RTE test accuracy the
entailment rule classifier was trained on the whole training set.

\subsubsection{Overall Lexical and Phrasal Entailment Evaluation}

Table~\ref{tab:evalall} shows the results of the Entailment experiments
on all rules, both lexical and phrasal. In order to give bounds on our system's
performance, we present baseline score (entailment rule classifier always 
predicts non-entailing) 
and ceiling score (entailment rule classifier always 
predicts gold standard annotation).

\begin{table}
\centering
 \footnotesize 
\begin{tabular}{|lrrr|}
    \hline
    {\bf Feature set} & {\bf Intrinsic} & {\bf RTE Train} & {\bf RTE Test}\\
    \hline
    Always guess neutral & 64.3 & 73.9 & 73.3  \\
    Gold standard annotations&100.0 & 95.0 & 95.5  \\
    \hline
    Base only            & 64.3 & 73.8 & 73.4  \\
    Wordform only        & 67.3 & 77.0 & 76.7  \\
    WordNet only         & 75.1 & 81.9 & 81.3  \\
    Dist (Lexical) only  & 71.5 & 78.7 & 77.7  \\
    Dist (Phrasal) only  & 66.9 & 75.9 & 75.1  \\
    Asym only            & 70.1 & 77.3 & 77.2  \\
    \hline
    All features         & 79.9 & 84.0 & 83.0  \\
    \hline
\end{tabular}
\caption{Cross-validation accuracy on Entailment on all rules}
\label{tab:evalall}
\end{table}

%\begin{table}
%\centering
% \footnotesize 
%\begin{tabular}{|p{3cm}|L|L|L|L|L|L|L|L|L|L|L|L|}
%    \hline
%    {\bf Feature set} & {\bf In} & {\bf Tr} & {\bf Ts} & {\bf In} & {\bf Tr} & {\bf Ts} & {\bf In} & {\bf Tr} & {\bf Ts} & {\bf In} & {\bf Tr} & {\bf Ts} \\
%    \hline
%    Always guess neutral & 64.3 & 73.9 & 73.3  \\
%    Gold standard annotations&100.0 & 95.0 & 95.5  \\
%    \hline
%    Base only            & 64.3 & 73.8 & 73.4  \\
%    Wordform only        & 67.3 & 77.0 & 76.7  \\
%    WordNet only         & 75.1 & 81.9 & 81.3  \\
%    Dist (Lexical) only  & 71.5 & 78.7 & 77.7  \\
%    Dist (Phrasal) only  & 66.9 & 75.9 & 75.1  \\
%    Asym only            & 70.1 & 77.3 & 77.2  \\
%    \hline
%    All features         & 79.9 & 84.0 & 83.0  \\
%    \hline
%\end{tabular}
%\caption{Cross-validation accuracy on Entailment on all rules}
%\label{tab:evalall}
%\end{table}
%

The ceiling score (entailment rule classifier always predicts gold
standard annotation) does not achieve perfect performance.  This is due to a
number of different issues including misparses, imperfect rules
generated by the modified Robinson resolution, a few system inference timeouts,
and various idiosyncrasies of the SICK dataset.

Another point to note is that WordNet is by far the strongest set of
features for the task. This is unsurprising, as synonymy and hypernymy
information from WordNet gives nearly perfect information for much of the task.
There are some exceptions, such as {\em woman~$\nrightarrow$~man}, or {\em
black~$\nrightarrow$~white}, which WordNet lists as antonyms, but
which are not
considered contradictions in the SICK dataset 
(e.g: ``T: A man is cutting a tomato'' and ``H: A woman is cutting a tomato'' is not a contradiction).
However, even though WordNet has extremely high coverage on this particular dataset, it
still is far from exhaustive: about a quarter of the rules have at least one
pair of words for which WordNet relations could not be determined.

The lexical distributional features do surprisingly well on
the task, obtaining a test accuracy of 77.7 (Table~\ref{tab:evalall}). This indicates that, even with {\em only} distributional similarity,
we do well enough to score in the upper half of systems in the original SemEval
shared task, where the median test accuracy of all teams was 77.1 \cite{marelli:semeval14}. Two components were critical to the
increased performance over our own prior work:
first, the use of multiple distributional spaces (one topical, one
syntactic); second, the binning of cosine values. While using only the
BoW cosine similarity as a feature, the classifier actually performs {\em below}
baseline (50.0 intrinsic accuracy; compare to Table~\ref{tab:evallexical}).
Similarly, only using syntactic cosine similarity as a feature also performs
poorly (47.2 IA). However adding binning to either improves performance (64.3 and
64.7 for BoW and Dep), and adding binning to both improves it further (68.8 IA,
as reported in Table~\ref{tab:evallexical}).

The phrasal distributional similarity features, which are based on the
state-of-the-art \namecite{paperno:acl14} compositional vector space, perform somewhat
disappointingly on the task.
We discuss possible reasons for this below in Section~\ref{subsubsec:phrasal}.

% One reason for this, is our implementation of
% \cite{paperno:acl14} only handles a limited number of syntactic operations
% and words, and has relatively low coverage of the dataset ($\sim94\%$).
% {\bf TODO: Pengxiang, can you think of any other reasons it doesn't
% so well?}
% KE: Why is 94% low coverage? Also I believe that our implementation
% of Paperno et al handles pretty much all the syntactic relations
% that the original paper handled. 

We also note that the Basic Alignment features and WordForm features
(described in Tables~\ref{tab:lexentfeat} and
\ref{tab:phrasalentfeat}) do not do
particularly well on their own. This is encouraging, as it means the dataset
cannot be handled by simply expecting the same words to appear on the LHS and
RHS. Finally, we note that the features are highly complementary, and the
combination of all features gives a substantial boost to performance.

\subsubsection{Evaluating the Lexical Entailment Rule Classifier}

Table~\ref{tab:evallexical} shows performance of the classifier on {\em only}
the lexical rules, which have single words on the LHS and RHS. In these
experiments we use the same procedure as before, but omit the phrasal rules
from the dataset. On the RTE tasks, we compute accuracy over only the SICK
pairs which require at least one lexical rule. Note that a new ceiling
score is needed, as some rules require both lexical and phrasal
predictions, but we do not predict any phrasal rules.

\begin{table}
\centering
 \footnotesize 
\begin{tabular}{|lrrr|}
    \hline
    {\bf Feature set} & {\bf Intrinsic} & {\bf RTE Train} & {\bf RTE Test}\\
    \hline
    Always guess neutral & 56.6 & 69.4 & 69.3 \\
    Gold standard annotations&100.0 & 93.2 & 94.6 \\
    \hline
    Wordform only        & 57.4 & 70.4 & 70.9 \\
    WordNet only         & 79.1 & 83.1 & 84.2 \\
    Dist (Lexical) only  & 68.8 & 76.3 & 76.7 \\
    Asym only            & 76.8 & 78.3 & 79.2 \\
    \hline
    All features         & 84.6 & 82.7 & 83.8 \\
    \hline
\end{tabular}
\caption{Cross-validation accuracy on Entailment on lexical rules only}
\label{tab:evallexical}
\end{table}

Again we see that WordNet features have the highest contribution. Distributional rules still perform better
than the baseline, but the gap between distributional features and
WordNet is much more apparent. 
Perhaps most encouraging is the very high performance of the Asymmetric features: by
themselves, they perform substantially better
than just the distributional features. We investigate this further
below in Section~\ref{subsubsec:asym}. 

As with the entire dataset, we once again see that all the features are
highly complementary, and intrinsic accuracy is greatly improved by
using all the features together. It may be surprising that these significant
gains in intrinsic accuracy do not translate to improvements on the
RTE tasks; in fact, there is a minor drop from using all features compared
to only using WordNet. This most likely depends on {\em which} pairs the
system gets right or wrong. For sentences involving multiple lexical rules,
errors become disproportionately costly. As such, the high-precision
WordNet predictions are slightly better on the RTE task.

In a qualitative analysis comparing a
classifier with only cosine distributional features to a classifier
with the full feature set, we found that, as expected, the distributional features
miss many hypernyms and falsely classify many co-hyponyms as
entailing: We manually analyzed a sample of 170 pairs that the distributional classifier
falsely classifies as entailing. Of these, 67 were co-hyponyms (39\%), 33 were
antonyms (19\%), and 32 were context-specific pairs like
\textit{stir/fry}. On the other hand, most (87\%)
 cases of entailment that the distributional classifier detects but the
all-features classifier misses are word pairs that have no link in
WordNet. These pairs include \textit{note
  $\to$ paper}, \textit{swimmer $\to$ racer}, \textit{eat $\to$ bite},
and \textit{stand $\to$ wait}.

\subsubsection{Evaluating the Phrasal Entailment Rule Classifier}
\label{subsubsec:phrasal}

Table~\ref{tab:evalphrasal} shows performance when looking at only the phrasal rules. As with the evaluation of lexical
rules, we evaluate the RTE tasks only on sentence pairs that use
phrasal rules, and do not provide any lexical inferences. As such, the
ceiling score must again be recomputed.

\begin{table}
\centering
 \footnotesize 
\begin{tabular}{|lrrr|}
    \hline
    {\bf Feature set} & {\bf Intrinsic} & {\bf RTE Train} & {\bf RTE Test}\\
    \hline
    Always guess neutral & 67.8 & 72.5 & 72.7  \\
    Gold standard annotations& 100.0& 91.9 & 92.8  \\
    \hline
    Base only            & 68.3 & 73.3 & 73.6  \\
    Wordform only        & 72.5 & 77.1 & 77.1  \\
    WordNet only         & 73.9 & 78.3 & 77.7  \\
    Dist (Lexical) only  & 72.9 & 77.0 & 76.5  \\
    Dist (Phrasal) only  & 71.9 & 75.7 & 75.3  \\
    \hline
    All features         & 77.8 & 79.7 & 78.8  \\
    \hline
\end{tabular}
\caption{Cross-validation accuracy on Entailment on phrasal rules only}
\label{tab:evalphrasal}
\end{table}

We first notice that the phrasal subset is generally harder than the lexical
subset: none of the features sets on their own provide dramatic improvements
over the baseline, or come particularly close to the ceiling score.
On the other hand, using all features together does better than
any of the feature groups by themselves, indicating again that the feature groups are highly
complementary.

Distributional features perform rather close to the Wordform features,
suggesting that possibly the Distributional features may simply be proxies for
the {\em same lemma} and {\em same POS} features. A qualitative analysis
comparing the predictions of Wordform and Distributional features shows
otherwise though: the Wordform features are best at correctly identifying
non-entailing phrases (higher precision), while the distributional features are
best at correctly identifying entailing phrases (higher recall). % Many of the
% entailments that the distributional features successfully identifies
% are complex paraphrases like {\em snow $\rightarrow$ snowy area},
% {\em toddler $\rightarrow$ young kid}, and {\em teenage $\rightarrow$ in teens}.

As with the full dataset, we see that the features based on \namecite{paperno:acl14}
do not perform as well as just the alignment-based distributional lexical features; in fact,
they do not perform even as well as features which make predictions using only
Wordform features. We qualitatively compare the Paperno et al.\
features (or phrasal features for short) to the features based on word similarity of
greedily aligned words (or alignment features). We generally find the phrase
features are much more likely to predict neutral, while the alignment-based features
are much more likely to predict entailing. In particular, the phrasal vectors
seem to be much better at capturing non-entailment based on
differences in prepositions 
({\em walk inside building $\nrightarrow$ walk outside building}),
additional modifiers on the RHS ({\em man $\nrightarrow$ old man},
{\em room $\nrightarrow$ darkened room}), and changing semantic roles
({\em man eats near kitten $\nrightarrow$ kitten eats}). Surprisingly,
we found the lexical distributional features were better at capturing
complex paraphrases, such as {\em teenage $\rightarrow$ in teens},
{\em ride bike $\rightarrow$ biker}, or {\em young lady $\rightarrow$
  teenage girl}.

\subsubsection{Evaluating the Asymmetric Classifier}
\label{subsubsec:asym}

\namecite{levy:2015tt} show several experiments suggesting that
asymmetric classifiers do not perform better at the task
of identifying hypernyms than when the RHS vectors alone are used as
features. That is, they find that the asymmetric classifier and
variants frequently learn to identify prototypical hypernyms rather
than the hypernymy relation itself. We look at our data in the light
of the Levy et al.\ study, in particular as none of the entailment
problem sets used by Levy et al.\ were derived from an existing RTE
dataset.

In a qualitative analysis comparing the predictions of a
classifier using only Asymmetric features with a classifier using only
cosine similarity, we found that the Asymmetric classifier does substantially
better at distinguishing hypernymy from co-hyponymy. This is what we
had hoped to find, as we had previously found an Asymmetric classifier to perform well at identifying hypernymy in other data
\cite{roller:coling14}, and cosine is known to heavily favor
co-hyponymy~\cite{baronilenci:gems2011}. However, we also find
that cosine
features are better at discovering synonymy, and that Asymmetric frequently
mistakes antonymy as an entailing.
We did a quantitative analysis comparing the predictions of a
classifier using only Asymmetric features to a classifier that tries
to learn typical hyponyms or hypernyms  by using only the LHS vectors,
or the RHS vectors, or both. Table~\ref{tab:evalasym} 
shows the results of these experiments. %We only consider
%rules with single words on the LHS and RHS, so baseline is the same as in Table~\ref{tab:evallexical}. 

\begin{table}
\centering
 \footnotesize 
\begin{tabular}{|lrrr|}
    \hline
    {\bf Feature set} & {\bf Intrinsic} & {\bf RTE Train} & {\bf RTE Test}\\
    \hline
    Always guess neutral & 56.6 & 69.4 & 69.3 \\
    Gold standard annotations&100.0 & 93.2 & 94.6 \\
    \hline
    Asym only            & 76.8 & 78.3 & 79.2 \\
    \hline
    LHS only             & 65.4 & 73.8 & 73.5 \\
    RHS only             & 73.2 & 78.6 & 79.9 \\
    LHS + RHS            & 76.4 & 79.8 & 80.6 \\
    \hline
    Asym + LHS + RHS     & 81.4 & 81.4 & 82.6 \\
    \hline
\end{tabular}
\caption{Cross-validation accuracy on Entailment on lexical rules for Asym evaluation}
\label{tab:evalasym}
\end{table}

Counter to the main findings of \namecite{levy:2015tt}, we find that
there is at least some learning of the entailment relationship by the
asymmetric classifier (in particular on the intrinsic evaluation), as
opposed to the prototypical hypernym hypothesis. We believe this is
because the dataset is too varied to allow the classifier to learn
what an entailing RHS looks like.  Indeed, a qualitative analysis
shows that the asymmetric features successfully predict many hypernyms
that RHS vectors miss. On the other hand, the RHS do manage to capture
particular semantic classes, especially on words that appear many
times in the dataset, like {\em cut, slice, man, cliff,} and {\em
  weight}.

The classifier given both the LHS and RHS vectors dramatically outperforms its
components: it is given freedom to nearly memorize rules that appear commonly
in the data.  Still, using all three sets of features (Asym + LHS + RHS) is
most powerful by a substantial margin. This feature set is able to capture the
frequently occurring items, while also allowing some power to generalize to
novel entailments. For example, by using all three we are able to capture some
additional hypernyms ({\em beer} $\rightarrow$ {\em drink}, {\em pistol}
$\rightarrow$ {\em gun}) and synonyms ({\em couch} $\rightarrow$ {\em sofa},
{\em throw} $\rightarrow$, {\em hurl}), as well as
some more difficult entailments ({\em hand} $\rightarrow$ {\em arm}, {\em young}
$\rightarrow$ {\em little}).

Still, there are many ways our lexical classifier could be improved, even using
all of the features in the system.  In
particular, it seems to do particularly bad on antonyms ({\em strike
$\nrightarrow$ miss}), and items that require additional world knowledge
({\em surfer $\rightarrow$ man}). It also occasionally misclassifies some
co-hyponyms ({\em trumpet $\nrightarrow$ guitar}) or gets the entailment
direction wrong ({\em toy $\nrightarrow$ ball}).

\subsection{RTE Task Evaluation}
\label{sec:rteeval}
This section evaluates different components of the system, and 
finds a configuration of our system that achieves
state-of-the-art results on the SICK RTE dataset. 

We evaluate the following system components. The component
\textbf{logic} is our basic MLN-based logic system that computes two
inference probabilities (Section~\ref{sec:hGivenT}). This includes the changes
to the logical form to handle the domain closure assumption
(Section~\ref{sec:dca}), the inference algorithm for query formulas
(Section~\ref{sec:qf}), 
and the inference optimization (Section~\ref{sec:mcw}).
The component \textbf{cws} deals with the problem that the
closed-world assumption raises for negation in the hypothesis
(Section~\ref{sec:cwa}), and \textbf{coref} is coreference resolution to
identify contradictions (Section~\ref{sec:coref}). The component
\textbf{multiparse} signals the use of two parsers, the top C\&C parse and the top
EasyCCG parse (Section~\ref{sec:boxer}).

The remaining components add entailment rules. The component
\textbf{eclassif} adds the rules from the best performing 
entailment rule classifier trained in 
Section~\ref{sec:lexeval}. 
This is the system with all features
included. 
The \textbf{ppdb} component adds rules from PPDB paraphrase
collection (Section~\ref{sec:ppdb}). 
The \textbf{wlearn} component learns a scaling factor for \textbf{ppdb} rules, 
and another scaling factor for the \textbf{eclassif} rules that maps
the classification confidence scores to MLN weights
(Section~\ref{sec:wlearn}).
% KE: I think we don't need to say more. Okay?
% Each rule has three classification confidence scores, one for each 
% rule type (Entail, Neutral or Contradict). 
% We use the rule type that 
% has the highest classification confidence score. 
Without weight learning, the scaling factor for \textbf{ppdb} is set to 1, 
and all \textbf{eclassif} rules are used as hard rules (infinite weight).
The \textbf{wlearn\_log}  component is similar to \textbf{wlearn} but uses 
equation \ref{eq:wlearnlog}, which first transforms a rule weight to
its log odds. 
The \textbf{wn} component adds rules from WordNet
(Section~\ref{sec:wn}). 
In addition, we have a few \textbf{handcoded}
rules (Section~\ref{sec:hand}). Like \textbf{wn}, the components \textbf{hyp} and
\textbf{mem} repeat information that is used as features for 
entailment rules classification but is not always picked up by the
classifier. As the classifier sometimes misses hypernyms, \textbf{hyp}
marks all hypernymy rules as entailing (so this component is subsumed by
\textbf{wn}), as well as all rules where the
left-hand side and the right-hand side are the same. (The latter step
becomes necessary after splitting long rules derived by our modified
Robinson resolution; some of the pieces may have equal left-hand and
right-hand sides.) The \textbf{mem} component memorizes all entailing rules seen in the training set of \textbf{eclassif}. 

Sometimes inference takes a long time, so we set a 2 minute timeout for each inference run. If inference 
does not finish processing within the time limit, we terminate the process and return an error code. 
About 1\% of the dataset times out. 

\begin{table}
\centering
 \footnotesize 
\begin{tabular}{|l@{\hspace{3pt}}c@{\hspace{3pt}}c|}
  \hline
  {\bf Components Enabled} & Train Acc. & Test Acc.\\
  \hline
%logic  &  63.16  &  63.45\\
% + cwa  &  72.12  &  71.71\\
% + cwa + coref  &  73.84  &  73.37\\
% + cwa + coref + ppdb  &  75.32  &  74.79\\
% + cwa + coref + ppdb + wlearn &  76.52  &  76.33\\
% + cwa + coref + ppdb + wlearn + wn  &  78.76  &  78.40\\
% + cwa + coref + ppdb + wlearn + wn + handcoded  &  79.2  &  78.79\\
% + cwa + coref + ppdb + wlearn + wn + handcoded + multiparse  &  80.84  &  \textbf{80.37}\\
logic  &  63.2  &  63.5\\
 + cwa  &  72.1  &  71.7\\
 + cwa + coref  &  73.8  &  73.4\\
 + cwa + coref + ppdb  &  75.3  &  74.8\\
 + cwa + coref + ppdb + wlearn &  76.5  &  76.3\\
 + cwa + coref + ppdb + wlearn + wn  &  78.8  &  78.4\\
 + cwa + coref + ppdb + wlearn + wn + handcoded  &  79.2  &  78.8\\
 + cwa + coref + ppdb + wlearn + wn + handcoded + multiparse  &  80.8  &  \textbf{80.4}\\
 \hline
\end{tabular}
\caption{Ablation experiment for the system components without \textbf{eclassif} }
\label{tab:nodist}
\end{table}

\subsubsection{Ablation Experiment without \textbf{eclassif}}
Because \textbf{eclassif} has the most impact on the system's accuracy, and when
enabled suppresses the contribution of the other components, 
we evaluate the other components first without \textbf{eclassif}. 
In the following section, we add the \textbf{eclassif} rules. 
Table \ref{tab:nodist} summarizes the results of this experiment.  The
results show that each component plays a role in improving the system
accuracy.  Our best accuracy without \textbf{eclassif} is 80.4\%.
Without handling the problem of negated hypotheses (\textbf{logic}
alone), $P(\lnot H|T)$ is almost always 1 and this additional
inference becomes useless, resulting in an inability to distinguish
between Neutral and Contradiction.  Adding \textbf{cwa}
significantly improves accuracy because the resulting system has 
$P(\lnot H|T)$ equal to 1 only for Contradictions.
 
Each rule set (\textbf{ppdb}, \textbf{wn}, \textbf{handcoded})
improves accuracy by reducing the number of false negatives. 
We also note that applying weight learning (\textbf{wlearn}) to find a
global scaling factor for PPDB
rules makes them more useful. 
The learned scaling factor is 3.0. When the knowledge base  is lacking
other sources, weight learning assigns a high scaling factor to PPDB,
giving it more influence throughout. When \textbf{eclassif} is added in
the following section, weight learning assigns PPDB a low scaling
factor because \textbf{eclassif} already
includes a large set of useful rules, such that only the highest weighted 
PPDB rules contribute significantly to the final inference. 

The last component tested is the use of  multiple parses (\textbf{multiparse}). 
Many of the false negatives are due to misparses. 
Using two different parses reduces the impact of the misparses, 
improving the system accuracy. 

\begin{table}
\centering
 \footnotesize 
\begin{tabular}{|l@{\hspace{3pt}}c@{\hspace{3pt}}c|}
  \hline
  {\bf Components Enabled} & Train Acc. & Test Acc.\\
  \hline
%logic + cwa + coref  & 73.84  &  73.37\\
%logic + cwa + coref + dist & 84.0 & 82.99\\
% + handcoded & 84.58 & 83.17\\
% + handcoded + multiparse & 85.02 & 83.86\\
%% + handcoded + multiparse + hyp & 85.30 & 84.03\\
% + handcoded + multiparse + hyp & 85.56 & 83.89\\
% + handcoded + multiparse + hyp + wlearn & 85.72 & 84.13\\
% + handcoded + multiparse + hyp + wlearn\_log & 85.90 & 84.27\\
% + handcoded + multiparse + hyp + wlearn\_log + mem & 93.42 & \textbf{85.06}\\
%% + handcoded + multiparse + hyp + mem & 90.88 & \textbf{84.43}\\
%% + handcoded + multiparse + hyp + mem + ppdb + wlearn (0.5) & 90.84 & 84.31 \\
%+ handcoded + multiparse + hyp + wlearn\_log + mem + ppdb & 93.36 & 84.94 \\
% \hline
% current state of the art \cite{lai:semeval14}  & -- & 84.58\\
logic + cwa + coref  & 73.8  &  73.4\\
logic + cwa + coref + eclassif & 84.0 & 83.0\\
 + handcoded & 84.6 & 83.2\\
 + handcoded + multiparse & 85.0 & 83.9\\
% + handcoded + multiparse + hyp & 85.30 & 84.03\\
 + handcoded + multiparse + hyp & 85.6 & 83.9\\
 + handcoded + multiparse + hyp + wlearn & 85.7 & 84.1\\
 + handcoded + multiparse + hyp + wlearn\_log & 85.9 & 84.3\\
 + handcoded + multiparse + hyp + wlearn\_log + mem & 93.4 & \textbf{85.1}\\
% + handcoded + multiparse + hyp + mem & 90.88 & \textbf{84.43}\\
% + handcoded + multiparse + hyp + mem + ppdb + wlearn (0.5) & 90.84 & 84.31 \\
+ handcoded + multiparse + hyp + wlearn\_log + mem + ppdb & 93.4 & 84.9 \\
 \hline
 current state of the art \cite{lai:semeval14}  & -- & 84.6\\
 \hline
\end{tabular}
\caption{Ablation experiment for the system components with \textbf{eclassif},
and the best performing configuration}
\label{tab:withdist}
\end{table}

\subsubsection{Ablation Experiment with \textbf{eclassif}}
In this experiment, we first use \textbf{eclassif} as a knowledge base, then incrementally add the 
other system components. Table \ref{tab:withdist} summarizes the results.
First, we note that adding \textbf{eclassif} to the knowledge base $KB$ significantly 
improves the accuracy from 73.4\% to 83.0\%. This is higher than
what \textbf{ppdb} and \textbf{wn} achieved without \textbf{eclassif}. 
Adding \textbf{handcoded} still improves the accuracy somewhat. 

Adding \textbf{multiparse} improves accuracy, but interestingly,
not as much as in the previous experiment (without \textbf{eclassif}). 
The improvement on the test set decreases from 1.6\% to 0.7\%. 
Therefore, the rules in \textbf{eclassif} help reduce the impact of misparses. 
Here is an  example to show how:
\textit{T: An ogre is jumping over a wall}, 
\textit{H: An ogre is jumping over the fence} which in logic are: 
\begin{itemize}
\item [$T$:] $\exists x,y,z. \: ogre(x) \wedge agent(y,x) \wedge jump(y) \wedge over(y,z) \wedge wall(z)$
\item [$H$:] $\exists x,y,z. \: ogre(x) \wedge agent(y,x) \wedge
  jump(y) \wedge over(y)  \wedge patient(y,z) \wedge wall(z)$
\end{itemize}
$T$ should entail $H$ (strictly speaking, \textit{wall} is not a \textit{fence} but this is a positive entailment example 
in SICK).
The modified Robinson resolution yields the following rule: 
\begin{itemize}
\item [$F$:] $ \forall x,y. \:  jump(x) \wedge over(x,y) \wedge wall(y) \Rightarrow 
jump(x) \wedge over(x)  \wedge patient(x,y) \wedge wall(y) $
\end{itemize}
Note that in $T$, the parser treats \textit{over} as a preposition, 
while in $H$, \textit{jump over} is treated as a particle verb. 
A lexical rule \textit{wall $\Rightarrow$ fence} is not enough to get the 
right inference because of this inconsistency in the parsing. 
The rule $F$ reflects this parsing 
inconsistency. When $F$ is translated to text for the entailment
classifier, we obtain 
\textit{jump over wall $\Rightarrow$ jump over fence}, which is a simple 
phrase that the entailment classifier addresses without dealing 
with the complexities of the logic. 
Without the modified Robinson resolution, we would have had to resort
to collecting ``structural'' inference rules like
$ \forall x,y. \:  over(x,y) \Rightarrow  over(x)  \wedge patient(x,y) $.

Table~\ref{tab:withdist} also shows the impact of \textbf{hyp} and \textbf{mem}, two
components that in principle should not add anything over
\textbf{eclassif}, but they do add some accuracy due to noise in the
training data of \textbf{eclassif}. 
% . As
% mentioned earlier, \textbf{hyp} and \textbf{mem} should have been
% captured by the lexical entailment rules classifier, and adding them should
% have not improved the accuracy of our system. However, the fact that
% they improve the accuracy is probably attributed to the inaccurate
% objective function used in the lexical entailment rules classifier as
% discussed in ... \textbf{TODO: This is a big issue that needs to be explained 
% much better than that. }

Weight learning results are the rows \textbf{wlearn} and \textbf{wlearn\_log}. 
Both weight learning components help improve the system's accuracy. 
It is interesting to see that even though the SICK dataset is not 
designed to evaluate "degree of entailment", it is still useful 
to keep the rules uncertain (as opposed to using hard rules)
and use probabilistic inference. 
Results also show that \textbf{wlearn\_log} performs slightly better than 
\textbf{wlearn}.

Finally, adding \textbf{ppdb} does not improve the
accuracy. Apparently, \textbf{eclassif} already captures all the useful rules
that we were getting from \textbf{ppdb}. It is interesting to see that
simple distributional information can subsume a large paraphrase database
like PPDB.  Adding \textbf{wn} (not shown in the table) leads to a
slight decrease in accuracy. 

The system comprising \textbf{logic}, \textbf{cwa}, \textbf{coref},
\textbf{multiparse}, \textbf{eclassif}, \textbf{handcoded},
\textbf{hyp}, \textbf{wlearn\_log}, and \textbf{mem} 
achieves a state-of-the-art accuracy score of 85.1\% on the
SICK test set. 
%reaching previously published state  of the art results \cite{lai:semeval14}. 
The entailment rule classifier
\textbf{eclassif} plays a vital role in this result.

\section{Future Work}
\label{sec:future}
One area to explore is contextualization. The evaluation of the 
entailment rule classifier showed that some of the entailments
are context-specific, like put/pour (which are entailing only for
liquids) or push/knock (which is entailing in the context of ``pushing
a toddler into a puddle''). Cosine-based distributional features were able to identify some of these cases
when all other features did not. We would like to explore whether \emph{contextualized}
distributional word representations, which take the sentence context
into account~\cite{erk:emnlp08,thater:acl10,Dinu:2012wi}, can identify
such context-specific lexical entailments more reliably. 
% The lexical
% substitution dataset of \namecite{Kremer:2014ww} will be a good resource
% for studying this question, as over 60\% of the substitutes in th

We would also like to explore new ways of measuring lexical
entailment. It is well-known that cosine
similarity gives  high ratings to
co-hyponyms~\cite{baronilenci:gems2011}, and our evaluation confirmed
that this is a problem for lexical entailment judgments, as
co-hyponyms are usually not entailing. However, co-hyponymy judgments
can be used to position unknown terms in the WordNet
hierarchy~\cite{SnowJurafskyNg:06}. This could be a new way of using
distributional information in lexical entailment: using cosine
similarity to position a term in an existing hierarchy, and then using
the relations in the hierarchy for lexical entailment. While
distributional similarity is usually used only on individual word pairs,
%as if nothing else was known about the language,
this technique would use distributional similarity to learn the meaning of unknown
terms given that many other terms are already known. 

While this paper has focused on the RTE task, we are interested in
applying our system to other tasks, in particular question answering task. 
%Question answering is 
%the task of finding an answer of a WH question from large text corpus.
This task is interesting because it may offer a
wider variety of tasks to the distributional subsystem.
%, including context-specific matches and the need to learn domain-specific distributional knowledge. 
Existing logic-based systems are usually applied to limited domains, 
such as querying a specific database~\cite{kwiatkowski:emnlp13,berant:emnlp13}, but with our system, we have the potential 
to query a large corpus because we are using Boxer for wide-coverage semantic analysis. 
The general system architecture discussed in this paper can be applied 
to the question answering task with some modifications. %One main
%modification is that % As in section \ref{sec:parse},
% all the text would be translated to logic using Boxer, then the quantifiers handling and CWA would be used without 
% change. However, 
%the task representation changes to finding the best entities in the text
%that fill in an existentially quantified variable of the query.
For knowledge base construction, the general idea of using theorem
proving to infer rules still applies, but the details of the technique 
would be a lot different from the Modified Robinson Resolution in section \ref{sec:rr}. 
For the probabilistic logic inference, scaling becomes a major
challenge. 

Another important extension to this work is to support generalized
quantifiers in probabilistic logic. Some determiners, such as ``few''
and ``most'', cannot be represented in standard first-order logic, and
are usually addressed using higher-order logics. But
it could be possible to represent them using the probabilistic aspect
of probabilistic logic, sidestepping the need for 
higher-order logic.

\section{Conclusion}
\label{sec:conclusion}
Being able to effectively represent natural language semantics is
important and has many important applications. We have introduced an
approach that uses probabilistic logic to combine the expressivity and
automated inference provided by logical representations, with the
ability to capture graded aspects of natural language captured by
distributional semantics.  We evaluated this semantic representation
on the RTE task which requires deep semantic understanding. Our system
maps natural-language sentences to logical formulas, uses them to build
probabilistic logic inference problems, builds a knowledge base from
precompiled resources and on-the-fly distributional resources, then
performs inference using Markov Logic.  Experiments demonstrated
state-of-the-art performance on the recently introduced SICK RTE task.

\begin{acknowledgments}
This research was supported by the DARPA DEFT program under AFRL grant
FA8750-13-2-0026, by the NSF CAREER grant IIS 0845925 and by the NSF
grant IIS 1523637.
Any opinions, findings, and conclusions or
recommendations expressed in this material are those of the author
and do not necessarily reflect the view of DARPA, DoD or the US government.
Some experiments were run on the Mastodon Cluster supported by NSF Grant
EIA-0303609, and the Texas Advanced Computing
Center (TACC) at The
  University of Texas at Austin. 
\end{acknowledgments}

\starttwocolumn

\bibliographystyle{fullname}
\bibliography{lunar-clean,additional-clean}

\end{document}